\PassOptionsToPackage{table,dvipsnames}{xcolor}
\documentclass{article}

    \PassOptionsToPackage{numbers, compress}{natbib}
 \usepackage[main, final]{neurips_2026}
\usepackage[utf8]{inputenc} % allow utf-8 input
\usepackage[T1]{fontenc}    % use 8-bit T1 fonts
\usepackage{hyperref}       % hyperlinks
\usepackage{url}            % simple URL typesetting
\usepackage{booktabs}       % professional-quality tables
\usepackage{amsfonts}       % blackboard math symbols
\usepackage{nicefrac}       % compact symbols for 1/2, etc.
\usepackage{microtype}      % microtypography
\usepackage{xcolor}         % colors

\usepackage{booktabs,tabularx,array,makecell}
\usepackage{booktabs,tabularx,array,makecell}

\usepackage{xcolor}
\usepackage{titlesec}
\usepackage{array}
\usepackage{relsize}
\usepackage{subcaption}
\usepackage{graphicx}
\usepackage{enumitem}
\usepackage{amsmath} 
\usepackage{amsthm} 
\usepackage{amssymb} 
\usepackage{mathtools}
\usepackage{wrapfig}
\usepackage{multirow}
\usepackage{booktabs}

\usepackage{tcolorbox}
\tcbuselibrary{skins,breakable}

    {\par\noindent\textbf{#1}\itshape} % Begin code
    {\par} % End code

\newtcolorbox{takeawaybox}{
    enhanced,
    breakable,
    colback=gray!10,  % Background color
    colframe=gray!50,  % Border color
    arc=2pt,          % Corner radius
    boxrule=1pt,      % Border thickness
    left=8pt,         % Left margin
    right=8pt,
    top=4pt,
    bottom=4pt,
    before upper={},  % Title
    fontupper=\normalfont,
    overlay unbroken={
        \node[anchor=north east, inner sep=4pt] at (frame.north east) {};
    }
}
\title{Unlocking the Duality between\\Flow and Field Matching}

\author{
  Daniil Shlenskii\\
  AXXX, Moscow, Russia\\
  Applied AI Institute, Moscow, Russia\\
  \texttt{daniil.shlenskii@gmail.com} \\
  \And
  Alexander Varlamov \\
  Applied AI Institute, Moscow, Russia\\
  \AND
  Nazar Buzun \\
  AXXX, Moscow, Russia \\
  Innopolis University, Innopolis, Russia \\
  \And
  Alexander Korotin\\
  Applied AI Institute, Moscow, Russia\\
  AXXX, Moscow, Russia\\
}

\definecolor{ifmcolor}{HTML}{009E73} % bluish green / teal
\definecolor{cfmcolor}{HTML}{D55E00} % vermilion / warm orange-red
\definecolor{conclusioncolor}{HTML}{6A7E39} % mixture

\colorlet{ifmBG}{ifmcolor!10!white} % E6F5F1
\colorlet{cfmBG}{cfmcolor!10!white} % FBEFE6
\colorlet{generalBG}{black!2}   % slightly cleaner than gray!7

\definecolor{condcolor}{HTML}{1F77B4}

\theoremstyle{plain}
\newtheorem{theorem}{Theorem}[section]
\newtheorem{proposition}[theorem]{Proposition}
\newtheorem{lemma}[theorem]{Lemma}
\newtheorem{corollary}[theorem]{Corollary}
\theoremstyle{definition}

\theoremstyle{remark}
\newtheorem{remark}[theorem]{Remark}

\newcommand{\cfmcolor}[1]{{\color{cfmcolor}\ifmmode#1\else\textbf{#1}\fi}}
\newcommand{\ifmcolor}[1]{{\color{ifmcolor}\ifmmode#1\else\textbf{#1}\fi}}
\newcommand{\condcolor}[1]{{\color{condcolor}\ifmmode#1\else#1\fi}}

\newcommand{\CFM}{{{CFM}}}
\newcommand{\IFM}{{{IFM}}}

\newcommand{\x}{\mathbf{x}}
\renewcommand{\v}{{\mathbf{v}}}
\renewcommand{\u}{{\mathbf{u}}}
\newcommand{\E}{{\mathbf{E}}}
\newcommand{\n}{\mathbf{n}}
\newcommand{\e}{\mathbf{\epsilon}}
\newcommand{\xt}{\tilde{\mathbf{x}}}

\newcommand{\cond}{{\x_0, \x_T}}
\newcommand{\ccond}{{\x_0{\color{condcolor}, \x_T}}}
\newcommand{\cjoint}{\pi_{0{\color{condcolor}, T}}}
\newcommand{\jointl}{{\pi_{0, L}}}
\newcommand{\jointt}{{\pi_{0, T}}}
\newcommand{\condl}{{\x_0, \x_L}}

\newcommand{\J}[1]{J(#1, \x_0)}
\newcommand{\Jdot}[1]{\dot{J}(#1, \x_0)}
\newcommand{\I}[1]{\mathcal{I}(#1, \cond)}
\newcommand{\Idot}[1]{\dot{\mathcal{I}}(#1, \cond)}
\newcommand{\s}[1]{\sigma(#1)}
\newcommand{\sdot}[1]{\dot{\sigma}(#1)}

\newcommand{\vcond}{{\v_t^\ccond}}
\newcommand{\pcond}{{p_t^\ccond}}
\newcommand{\Econd}{{\E^\condl}}

\newcommand{\Eccondt}{{\E^\ccond}}

\newcommand{\rom}[1]{\uppercase\expandafter{\romannumeral #1}}

\newcommand{\Id}{I}

\newcommand{\real}{\mathbb{R}}
\renewcommand{\dim}{D}
\newcommand{\extradim}{d}
\newcommand{\batchsize}{N}

\newcommand{\new}[1]{\textcolor{black}{#1}}
\newcommand{\neww}[1]{\textcolor{black}{#1}}
\begin{document}
\maketitle
\begin{abstract}
% Conditional Flow Matching (CFM) unifies conventional generative paradigms such as diffusion and flow-based models. Interaction Field Matching (IFM) is a recently proposed framework that generalizes Electrostatic Field Matching (EFM), rooted in Poisson Flow Generative Models (PFGM). While both frameworks define generative dynamics, they start from different objects: CFM specifies a conditional probability path in data space, whereas IFM specifies a physics-inspired interaction field in an augmented data space.
% This raises a basic question: \textbf{are CFM and IFM genuinely different, or are they two descriptions of the same underlying dynamics?} We show that they coincide for a natural subclass of IFM, which we call \textit{forward-only} IFM. Specifically, we construct mappings between CFM and forward-only IFM and show that they induce the same generative dynamics.
% We further show that general IFM is strictly more expressive: it includes EFM and other interaction fields that cannot be realized within the standard CFM formulation.
% Our findings suggest developing generative models using both interpretations rather than treating them separately. Moreover, they highlight a novel direction for generative modeling based on backward-oriented field lines, which lies outside the conventional CFM formalism and may lead to new generative properties.

Conditional Flow Matching (CFM) unifies conventional generative paradigms such as diffusion and flow-based models. Interaction Field Matching (IFM) is a recently proposed framework that generalizes Electrostatic Field Matching (EFM), rooted in Poisson Flow Generative Models (PFGM). While both frameworks define generative dynamics, they start from different objects: CFM specifies a conditional probability path in data space, whereas IFM specifies a physics-inspired interaction field in an augmented data space.
This raises a basic question: \textbf{are CFM and IFM genuinely different, or are they two descriptions of the same underlying dynamics?}
We show that they coincide for a natural subclass of IFM, which we call \textit{forward-only} IFM.
\new{Specifically, we give explicit constructions in both directions and
show that CFM and forward-only IFM induce the same family of data-space
dynamics.
General IFM, however, is strictly more expressive: already in EFM, its
sampling rule yields transport maps that reverse orientation 
with positive probability,
something no CFM flow can do.
Our findings suggest using the probabilistic and geometric views jointly
when developing generative models, and exploring field-based transport
beyond the forward-only regime.}\\
Code: {\footnotesize\url{https://github.com/daniil-shlenskii/flow-field-matching-duality}}.

% Conditional Flow Matching (CFM) unifies conventional generative paradigms such as diffusion models and flow matching. Interaction Field Matching (IFM) is a newer framework that generalizes Electrostatic Field Matching (EFM) rooted in Poisson Flow Generative Models (PFGM). While both frameworks define generative dynamics, they start from different objects: CFM specifies a conditional probability path in data space, whereas IFM specifies a physics-inspired interaction field in an augmented data space.
% This raises a basic question: \textbf{are CFM and IFM genuinely different, or are they two descriptions of the same underlying dynamics?} We show that they coincide for a natural subclass of IFM that we call forward-only IFM. Specifically, we construct a bijection between CFM and forward-only IFM. We further show that general IFM is strictly more expressive: it includes EFM and other interaction fields that cannot be realized within the standard CFM formulation. 
% Our findings suggest to develop generative model using both interpretations rather than develop them separately, and, moreover, emphasise a novel direction of generative models with backward oriented-lines, which does not follow conventional CFM formalism so may inherit some novel generative properties.

% Finally, we highlight how this duality can benefit both frameworks: it provides a probabilistic interpretation of forward-only IFM and yields novel, IFM-driven techniques for CFM.
\end{abstract}

\begin{figure}[h]
    \vspace{-20pt}
    \centering
    \includegraphics[width=\textwidth]{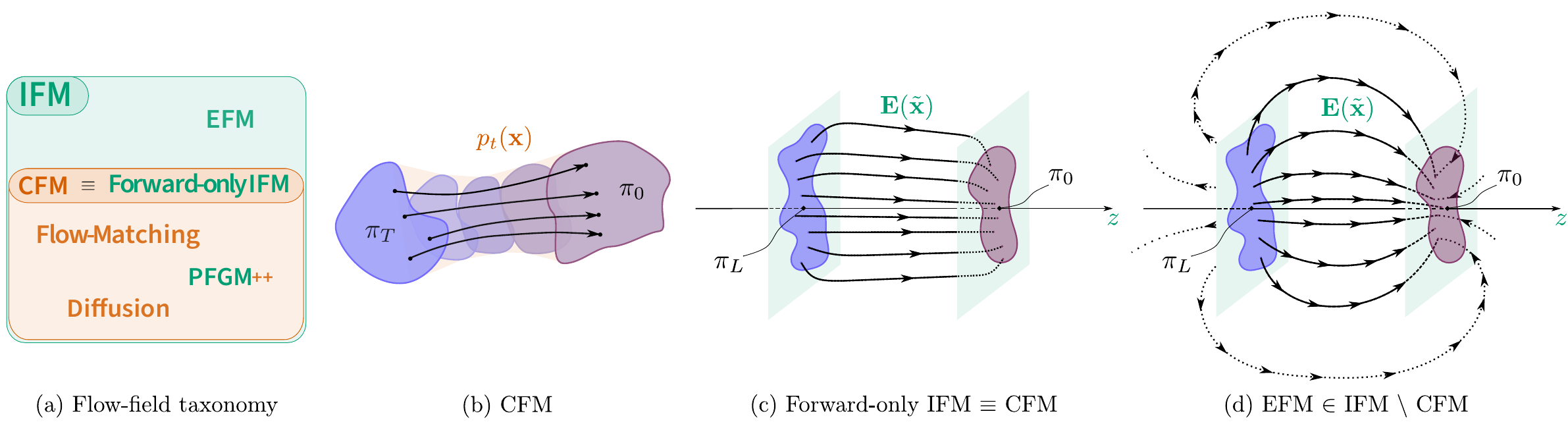}
    % \vspace{-15pt}
    \caption{
        \textbf{Overview of the flow--field duality.}
        We establish that CFM and forward-only IFM induce the same data-space generative dynamics (\S\ref{sec:duality}):
        $\mathrm{CFM} \equiv \mathrm{forward\text{-}only\ IFM} \subsetneq \mathrm{IFM}$.
        Panel (a) summarizes the resulting taxonomy.
        While conventional CFM (b, \S\ref{sec:conditional_flow_matching}) admits an equivalent forward-only field representation (c, \S\ref{sec:interaction_field_matching}),
        \new{general IFM is strictly more expressive: it includes EFM
        (d, \S\ref{sec:interaction_field_matching}), which admits returning trajectories
        whose transport maps reverse orientation and therefore cannot be
        realized by any CFM flow (\S\ref{sec:beyond_foifm}).}
    }
    \label{fig:teaser}
    \vspace{-10pt}
\end{figure}

\section{Introduction}\label{sec:introduction}
Modern continuous-time generative modeling is largely organized around flow-based descriptions of transport. 
Diffusion~\citep{sohl2015deep,song2020score} and flow-based~\citep{liu2022flow,lipman2022flow,albergo2022building} generative models can both be unified under the Conditional Flow Matching~\citep[CFM]{tong2023improving} framework. 
CFM constructs a generative ODE by specifying a conditional probability path in data space together with a compatible conditional velocity field, and then marginalizing these conditional objects to obtain global dynamics that transport the source distribution to the target distribution (Figure~\ref{fig:teaser}b).

A separate line of work has developed physics-inspired~\citep{liu2023genphys,shlenskii2025overclocking}, field-based generative models, including Poisson Flow Generative Models~\citep[PFGM]{pfgm}, PFGM++~\citep{pfgm++}, and Electrostatic Field Matching~\citep[EFM]{kolesov2025field}. 
Rather than starting from probability paths in data space, these methods define vector fields in an augmented space and generate samples by following their field lines. 
This perspective is formalized by the recently proposed Interaction Field Matching~\citep[IFM]{manukhov2025interaction} framework, which constructs particle-dependent interaction fields and combines them via a superposition principle to obtain a global transport field (Figure~\ref{fig:teaser}c,d).

Thus, CFM and IFM define generative dynamics from different first principles: CFM is path-first and probabilistic, whereas IFM is field-first and geometric. 
Understanding whether these views are genuinely distinct is important because it determines whether techniques, objectives, and intuitions developed in one framework can be transferred to the other. 
This raises a basic question: \textit{are CFM and IFM fundamentally distinct, or do they produce the same generative dynamics expressed in different languages?}
\textbf{Our main contributions} are as follows:

\begin{itemize}[leftmargin=*, align=left]
    \item \textbf{Flow--field duality.}
    \new{We establish the equivalence between CFM and forward-only IFM,
    a natural subclass of IFM (\S\ref{sec:interaction_field_matching}),
    through explicit constructions in both directions
    (\S\ref{sec:from_flows_to_fields},
    \S\ref{sec:from_fields_to_flows}).
    These constructions preserve the induced data-space dynamics,
    providing probabilistic and geometric descriptions of the
    same generative process.}

    \item \textbf{A strict separation between CFM and general IFM.}
    \new{We show that general IFM is strictly more expressive than CFM
    at the level of data-space transport maps (\S\ref{sec:beyond_foifm}):
    while CFM transport maps never reverse orientation, general IFM admits maps that do.
    We also show that EFM is a particular instance of IFM realizing this property.}
\end{itemize}

These findings are important for two reasons. First, they suggest
developing generative models using the probabilistic and geometric
interpretations jointly, rather than treating them as separate
frameworks. 
Second, \new{they identify orientation reversal as a concrete mechanism
by which general IFM goes beyond CFM, motivating field-based generative
modeling outside the forward-only regime as a new research direction.}
\section{Preliminaries: CFM vs. IFM?}\label{sec:background}
In this section, we review the two constructions of generative dynamics that
will be compared throughout the paper. Both frameworks define \new{ODE-based} dynamics, which transport a source/prior distribution,
denoted by $\pi_T$ or $\pi_L$ depending on the framework's notation, to a
target distribution $\pi_0$.
First, in \S\ref{sec:conditional_flow_matching}, we recall CFM, a well-established framework that
generalizes diffusion and flow-based generative models~\citep{gao2025diffusion}.
Next, in \S\ref{sec:interaction_field_matching}, we review the recently
proposed IFM,
which can be viewed as a generalization of electrostatic generative models.
In \S\ref{sec:cfm_vs_ifm}, we provide a side-by-side comparison of the key
ingredients of the two approaches. Finally, in \S\ref{sec:pfgm}, we discuss
PFGM, which, to our knowledge,
is the only existing framework that has been found to admit both flow and field descriptions.
Throughout this section, we use \cfmcolor{orange} and \ifmcolor{green} to
highlight terms specific to CFM and IFM, respectively.

\subsection{Conditional Flow Matching (CFM)}
\label{sec:conditional_flow_matching}
\textbf{Intuition.}
The CFM framework begins by specifying two conditional objects in the data
space $\real^\dim$:
\textbf{(a)} a \cfmcolor{conditional distribution path} $\pcond$ that
interpolates between $\x_0 \in \real^\dim$ and
\condcolor{$\x_T \in \real^\dim$}, i.e.,
$p_{t=0}^\cond = \delta_{\x_0}$ and
\condcolor{$p_{t=T}^\cond = \delta_{\x_T}$}; and
\textbf{(b)} a compatible \cfmcolor{conditional velocity}
$\vcond(\x): \real^\dim \to \real^\dim$ associated with this path.
In the two-sided setting, both objects are conditioned on a target--source sample
pair $\ccond$.
Marginalizing these conditional objects with respect to a given coupling
$\cjoint(\ccond)$ yields a global velocity $\v_t$, which induces the
generative dynamics
${d\x} = \v_t(\x)\,{dt}$,
used to transport samples from the source distribution $\pi_T$ to the target
distribution $\pi_0$.

We use \condcolor{blue} to emphasize that the elementary \CFM{} objects
$\pcond$ and $\vcond$ can be constructed in either a
\textit{one-sided} or \textit{two-sided} manner.
In the one-sided case, $p_t^{\x_0}$ and $\v_t^{\x_0}$ are defined using
only target samples $\x_0 \sim \pi_0$.
In the two-sided case, $p_t^{\cond}$ and $\v_t^{\cond}$ are defined from
paired target and source samples $\cond \sim \jointt$, for example under the
independent coupling $\jointt = \pi_0 \times \pi_T$.
This choice does not affect the theoretical guarantees~\citep{tong2023improving}.
The one-sided setting is common in noise-to-data generation
~\citep{ho2020denoising}, where the source distribution $\pi_T$ is a simple
prior, such as a standard Gaussian, independent of the target distribution
$\pi_0$. By contrast, the two-sided setting is more typical for data-to-data
translation~\citep{liu2022flow,albergo_data_dependent}.

\textbf{Definition (Conditional Flow).}
A \emph{Conditional Flow}~\citep{tong2023improving} is specified by a pair $(\vcond(\x),$ $\pcond(\x))$ consisting of a conditional velocity field and a path of conditional distributions.
Both objects depend on $\ccond$ and satisfy the \cfmcolor{\textit{continuity equation}} $\forall \x \in \mathbb{R}^{\dim}$ and $t \in (0, T)$
\begin{align}\label{eq:cfm_cond_transport_equation}
        \partial_t p_t^\ccond(\x)
        + \nabla_{\x} \cdot \big( \v_t^\ccond(\x)\, p_t^\ccond(\x) \big)
        = 0,
\end{align}
where $\nabla_{\x}\cdot$ denotes the divergence operator with respect to $\x$,
together with the \textit{boundary conditions}
\begin{align}\label{eq:cfm_cond_boundaries}
    \lim_{t \to 0} p^\ccond_{t}(\x) = \delta_{\x_0}(\x), \,
    {\color{condcolor} \lim_{t \to T} p^\ccond_{t}(\x) = \delta_{\x_T}(\x)},
\end{align}
that hold in the sense of weak convergence.

\textbf{Generative dynamics.}
This pointwise-defined conditional flow induces a generative model that transports a source distribution $\pi_T(\x)$ to a target distribution $\pi_0(\x)$.
First, marginalizing the conditional path with respect to the data distribution $\cjoint$
yields a \cfmcolor{path of marginal distributions}
\begin{align}\label{eq:cfm_marginals}
    {p_t}(\x)
    = \int \pcond(\x)\, \pi_{0\condcolor{,T}}(\ccond)\, d\x_0\, \condcolor{d\x_T},
\end{align}
which satisfies $p_{t=0} = \pi_0$ and $p_{t=T} = \pi_T$.
Second, one can show that the marginals $p_t$ are generated by a deterministic flow governed by the ODE
\begin{align}\label{eq:cfm_ode_and_drift}
    {d\x}/{dt}
    =
    \v_t(\x)
    :=
    \mathbb{E}_{p_t(\ccond \mid \x_t = \x)}
        \vcond(\x),
\end{align}
where the \cfmcolor{global velocity} $\v_t(\x)$ is the conditional
expectation of the conditional velocity $\vcond(\x)$ over endpoint pairs
$\ccond$, given the intermediate state $\x_t=\x$.

Although the drift expression~\eqref{eq:cfm_ode_and_drift} is explicit, it is generally intractable to evaluate in practice, since sampling from the conditional distribution $p_t(\ccond \mid \x_t)$ is infeasible.

\textbf{Training.}
The conditional expectation in~\eqref{eq:cfm_ode_and_drift} naturally yields a regression objective for estimating the drift.
In particular, $\v_t(\x)$ can be learned by minimizing
\begin{align}\label{eq:cfm_objective}
    {
        \mathbb{E}_{t, \ccond,\, \x_t}
        \left\|
            \v_t(\x_t) - \vcond(\x_t)
        \right\|_2^2,
    }
\end{align}
where $t\sim\mathcal{U}[0, T]$, $\x_t \sim \pcond$, and the conditional $\vcond$ provides a \cfmcolor{single-sample target}.

\phantomsection
\label{sec_extra:two_sided_stochastic_interpolant}
\textbf{Instantiation~\rom{1} (Two-sided Stochastic Interpolant).}
One of the most common instances of the CFM is the two-sided Stochastic Interpolant~\citep[\S 2.1]{albergo2023stochastic}, defined by $\x_t = \I{t} + \s{t}\,\e$, where
\textbf{(a)} $\I{0} = \x_0$ and $\I{T} = \x_T$;
\textbf{(b)} $\s{0} = 0$, $\s{T} = 0$, and $\s{t} > 0$ for all $t \in (0, T)$;
\textbf{(c)} $\e \sim \mathcal{N}(0, \Id)$ is independent of $\cond$.
This construction induces a path of conditional distributions $p_t^\cond(\x) = \mathcal{N}\big(\I{t}, \s{t}^2 \Id\big)$, with the associated conditional velocity
$\v_t^\cond(\x_t) = \Idot{t} + ({\sdot{t}}/{\s{t}}) \big(\x_t - \I{t}\big)$,
where a dot over a function denotes the partial derivative with respect to time.

\phantomsection
\label{sec_extra:one_sided_stochastic_interpolant}
\textbf{Instantiation~\rom{2} (One-sided Stochastic Interpolant).}
Another commonly used instance is the one-sided Stochastic Interpolant~\citep[\S 3.2]{albergo2023stochastic}, also known as Flow Matching~\citep{lipman2022flow}, defined by $\x_t = \J{t} + \s{t}\,\e$, where
\textbf{(a)} $\J{0} = \x_0$ and $\J{T} = \mathbf{0}$;
\textbf{(b)} $\s{0} = 0$, $\s{T} = 1$, and $\s{t} > 0$ for all $t \in (0, T)$;
\textbf{(c)} $\x_T \sim \mathcal{N}(0, \Id)$ is independent of $\x_0$.
As in the two-sided case, this yields a path of conditional distributions
$p_t^{\x_0}(\x) = \mathcal{N}\big(\J{t}, \s{t}^2 \Id\big)$, with the corresponding conditional velocity
$\v_t^{\x_0}(\x_t) = \Jdot{t} + ({\sdot{t}}/{\s{t}}) \big(\x_t - \J{t}\big)$.

\textit{Remark.} Although this construction conditions only on $\x_0$, it still specifies a right-endpoint prior $\pi_T$ that is independent of $\pi_0$, namely the standard Gaussian:
$\pi_T(\x_T \mid \x_0) = \pi_T(\x_T) = \mathcal{N}(0, \Id)$.
Moreover, in the special case $\J{t} = \x_0 (1 - t/T)$ and $\s{t} = t/T$, one recovers the classical linear interpolant introduced in~\citep{lipman2022flow}.
The same choice is also closely related to the Gaussian flow-matching
formulation of diffusion models~\citep{gao2025diffusion}.

\subsection{Interaction Field Matching (IFM)}\label{sec:interaction_field_matching}
\textbf{Intuition.}
Interaction Field Matching (IFM) lifts the data space to an extended space
$\real^{\dim+1}$. Each data point $\x\in\real^\dim$ is augmented with an
additional coordinate $z\in\real$, giving $\xt=(\x,z)$. The target and source
distributions are placed on two parallel plates: the target distribution
$\pi_0$ lies on the plate $z=0$, while the source distribution $\pi_L$ lies on
the plate $z=L>0$. Thus, samples $\x_0\sim\pi_0$ and $\x_L\sim\pi_L$ are
identified with the augmented points $\xt_0=(\x_0,0)$ and $\xt_L=(\x_L,L)$.
In EFM, this geometry
is interpreted as a $(\dim+1)$-dimensional electric capacitor.

The elementary object in IFM is an \ifmcolor{interaction field}
$\E^\condl(\xt):\real^{\dim+1}\to\real^{\dim+1}$, defined for an endpoint pair
$\condl=(\x_0,\x_L)$. Its field lines connect the two augmented endpoints
$\xt_0$ and $\xt_L$. A \ifmcolor{global field} $\E(\xt)$ is then obtained by
\ifmcolor{superposition}: the interaction fields are averaged over endpoint
pairs $(\x_0,\x_L)\sim\jointl$.
The field lines of the resulting global field connect the two plates and,
traversed from $z=L$ to $z=0$, define the transport from the source $\pi_L$ to the target $\pi_0$.
% \textbf{Intuition.}
% The IFM framework operates in an \ifmcolor{extended space} $\real^{\dim+1}$, where each data point $\x\in\real^\dim$ is augmented with an additional coordinate $z\in\real$ to form $\xt=(\x,z)$.
% The target and source distributions are placed on two separate \textit{plates} located at $z=0$ and $z=L>0$, respectively.
% Accordingly, samples $\x_0\sim\pi_0$ and $\x_L\sim\pi_L$ correspond to $\xt_0=(\x_0,0)$ and $\xt_L=(\x_L,L)$.
% In Electrostatic Field Matching~\citep[EFM]{kolesov2025field}, this geometric setup is interpreted as a $(\dim+1)$-dimensional \emph{electric capacitor}.

% Within this construction, IFM begins by introducing an \ifmcolor{interaction field}
% $\E^\condl(\xt): \real^{\dim+1}\to\real^{\dim+1}$,
% defined for a pair of points $\condl$, whose \ifmcolor{field lines} connect $\xt_0$ and $\xt_L$.
% A corresponding \ifmcolor{global field} $\E(\xt)$ is then obtained via the \ifmcolor{superposition principle} by averaging these interaction fields over a given joint distribution $\jointl$.
% The field lines of the resulting global field define the transport from the source distribution $\pi_L$ to the target distribution $\pi_0$.

\textit{Remark.}
In contrast to conditional flows, interaction fields are inherently two-sided: their field lines are required to connect two particles located at $\xt_0$ and $\xt_L$.
This is reflected in our notation: for conditional flows, the source conditioning point $\condcolor{\x_T}$ is shown in \condcolor{blue}, whereas for interaction fields the source point $\x_L$ is treated on the same basis as the target point $\x_0$ and is shown in black.

\textbf{Definition (Interaction Field).} 
A vector field $\Econd(\xt) \in \real^{\dim + 1}$
is an \emph{Interaction Field}~\citep{manukhov2025interaction}
if it satisfies the following three properties.

\new{First}, for any closed surface $\partial M$ enclosing $\xt_0$ but not $\xt_L$, the total flux
\begin{align}\label{eq:ifm_total_flux}
    \int_{\partial M} \E^{\condl}(\xt)\cdot \mathbf{dS} \;=\; \Phi_0
\end{align}
is independent of the particle locations $\xt_0, \xt_L$, i.e., the field has a \ifmcolor{\textit{fixed total flux}}.

\new{Second}, the field is \textit{\ifmcolor{divergence-free (flux-conserving)}}:
\begin{align}\label{eq:if_conserves_flux}
    \nabla \cdot \Econd(\xt) = 0,
    \quad
    \forall
    \xt \in \real^{\dim + 1} 
        \backslash \{\xt_0, \xt_L \}
\end{align}

\new{Third, all \ifmcolor{field lines}, except for a family carrying zero
flux through every transversal hypersurface
\footnote{
\new{A hypersurface $S \subset \real^{\dim+1}$ is called \emph{transversal} if
$\Econd \cdot \mathbf{n} > 0$ along it, where $\mathbf{n}$ is a unit normal to $S$.}
},
connect the two particles:}
\begin{align}\label{eq:if_connects_particles}
    {d\xt(\tau)}/{d\tau} = \Econd(\xt(\tau)), 
    \quad
    \lim_{\tau\downarrow \tau_s} \xt(\tau) = \xt_0,
    \quad
    \lim_{\tau\uparrow \tau_f} \xt(\tau) = \xt_L,
\end{align}
where $\tau_s$ and $\tau_f$ denote the start and end values of the field-line parameter, respectively.

\textit{Remark.} The parameter $\tau$ in~\eqref{eq:if_connects_particles} is not a global schedule like the time variable in Conditional Flows~\eqref{eq:cfm_cond_transport_equation}.
Rather, $\tau$ is an arbitrary scalar parameter used solely to trace a field line (integral curve) of $\Econd(\xt)$.

\textbf{Special case (forward-only Interaction Field).}
We say that an interaction field is \textit{forward-only} if \textbf{(a)} it is defined in the region between the plates ($0 \le z \le L$), and, \textbf{(b)} its $z$-component satisfies $\Econd(\xt)_z>0$ everywhere the field is nonzero.

\textbf{Generative dynamics.}
Given the target and source distributions $\pi_0$ and $\pi_L$, the induced \ifmcolor{global field} $\E(\xt)$ is defined via a \ifmcolor{\emph{generalized superposition principle}} as
\begin{align}\label{eq:field_full}
    \E(\xt)
    = \int \Econd(\xt)\, \pi_{0,L}(\condl)\, d\x_0\, d\x_L.
\end{align}
That is, $\E(\xt)$ is the average contribution of interaction fields over particle pairs.
As shown in~\citep{manukhov2025interaction}, one can transport samples from $\pi_L$ to $\pi_0$ following the field lines of $\E$:
\begin{align}\label{eq:ifm_field_lines}
    {d\xt(\tau)}/{d\tau} = \E(\xt(\tau)).
\end{align}

\textit{In general,} simulating~\eqref{eq:ifm_field_lines} poses practical difficulties:
\textbf{(a)} field lines may move both forward ($\E(\xt)_z > 0$) and backward ($\E(\xt)_z < 0$) along the $z$-axis; and
\textbf{(b)} \new{field lines may leave the capacitor region $0\le z\le L$ and cross the plates several times, which requires the stochastic starting and stopping rules of~\citet[Appendix~A.2]{manukhov2025interaction}.}
\textit{For a forward-only interaction field,} the global field is forward-only as well, i.e., $\E_z>0$, and~\eqref{eq:ifm_field_lines} can be reparameterized using the physically meaningful coordinate $z$:
\begin{align}\label{eq:ifm_field_lines_z}
    {d\xt}/{dz}
    =
    {d\xt}/{d\tau}\cdot{d\tau}/{dz}
    =
    \E(\xt) \cdot \E(\xt)_z^{-1}
    =
    \left(
        {\E(\xt)_\x} / {\E(\xt)_z},\,
        1
    \right).
\end{align}
This form yields explicit boundaries, $z(\tau_s)=0$ and $z(\tau_f)=L$, and induces generative dynamics
\begin{align}\label{eq:ifm_sampling}
    {d\x}/{dz}
    =
    {\E(\xt)_\x}/{\E(\xt)_z},
\end{align}
which
is defined in the data space $\real^{\dim}$
and
$z$ here plays the role of a time-like variable.

\textbf{Training.}
For a general Interaction Field, one has to estimate the full field $\E(\xt) \in \real^{\dim+1}$ at every point $\xt \in \real^{\dim+1}$ where the field is defined.
For forward-only fields, however, it suffices to estimate the \emph{normalized} field ${\E(\xt)}/{\|\E(\xt)\|_2}$.
Indeed, the generative dynamics~\eqref{eq:ifm_sampling} can be rewritten as
\begin{align*}
    {d\x}/{dz}
    =
    {\E(\xt)_\x}/{\E(\xt)_z}
    =
    {\E(\xt)_\x}/{\|\E(\xt)\|_2}
    \cdot
    {\|\E(\xt)\|_2}/{\E(\xt)_z}.
\end{align*}

Motivated by this observation, \citet{manukhov2025interaction} use the following training objective:
\begin{align}\label{eq:ifm_objective}
    \mathbb{E}_{\xt \sim p_{\mathrm{vol}}}
    \left\|
        f_\theta(\xt)
        -
        {\E(\xt)}/{\|\E(\xt)\|_2}
    \right\|_2^2,
    \quad
    \E(\xt) = (\E(\xt)_\x, \E(\xt)_z)
\end{align}
where
\textbf{(a)} $f_\theta : \real^{\dim+1} \to \real^{\dim+1}$ is a neural network that estimates the normalized Interaction Field;
\textbf{(b)} $p_{\mathrm{vol}}$ is a heuristic \ifmcolor{distribution intended to cover the volume} between the plates at $z=0$ and $z=L$, where the field is defined; and
\textbf{(c)} $\E(\xt)$ is the global field estimated from data samples according to the generalized superposition principle~\eqref{eq:field_full}.

% This objective has two main drawbacks.
% First, it is unclear how to choose an appropriate volume-covering distribution $p_{\mathrm{vol}}$ for a given Interaction Field $\E(\xt)$.
% Second, estimating the target $\E(\xt)$ typically requires many data samples to reduce bias.

% This stands in contrast to Conditional Flow Matching, where the effective training volume is defined implicitly by the conditional density path $p_t^{\ccond}$, and the training target is determined from a single data pair.

% That said, the IFM objective is expected to exhibit lower variance than CFM, since its multi-sample target estimator typically has reduced variance compared to the single-sample targets used in CFM.

\textbf{Instantiation~\rom{1} (Electrostatic Field).}
An early example of an interaction field is provided by the EFM.
Consider a unit point charge of sign $\pm 1$ located at $\xt' \in \real^{\dim+1}$.
By Coulomb's law, it induces the field
$\E^{\xt'}_\pm(\xt) \propto \pm{(\xt - \xt')}/{\|\xt - \xt'\|_2^{\dim + 1}}$.
Assigning a positive unit charge to augmented samples $\xt_0 \sim \pi_0 \times \delta_{z=0}$ and a negative to $\xt_L \sim \pi_L \times \delta_{z=L}$ yields, by the electrostatic superposition principle~\citep{pfgm,kolesov2025field}, the pairwise-defined field
$\Econd(\xt) = \E^{\xt_0}_{+}(\xt) + \E^{\xt_L}_{-}(\xt)$.
\new{This field is an interaction field in the sense defined above.
Moreover, its field lines may move backward along the $z$-axis and leave the
capacitor region $0 \le z \le L$; the EFM field is therefore not forward-only,
and sampling uses the general IFM stochastic rule.}

\phantomsection
\label{sec_extra:original_ifm_realization}
\textbf{Instantiation~\rom{2} (IFM Field Realization).}
As discussed above, fields with backward-oriented lines are not straightforward to use in practice.
To address this, the authors of  IFM propose a forward-only realization.
They decompose the field as $\Econd(\xt)=\|\Econd(\xt)\|_2\,\mathbf{n^\condl}(\xt)$, where $\xt=(\x,z)$.
In~\citep[Appendix A.4]{manukhov2025interaction}, they give a concrete procedure to specify $\|\Econd(\xt)\|_2$ and $\mathbf{n^\condl}(\xt)$, which we reproduce in Appendix~\ref{app:original_ifm_field} for convenience.

\subsection{CFM vs.\ IFM}
\label{sec:cfm_vs_ifm}
Previously, we reviewed the CFM and IFM frameworks separately.
Although their differences may already be apparent, we provide \textbf{Table~\ref{tab:cfm_vs_ifm}} in
Appendix~\ref{app:cfm_vs_ifm} for convenience; it lists the main terms of both frameworks side by side.
The key differences can be summarized as follows.
First, the IFM \ifmcolor{global field} can be estimated by Monte Carlo averaging over particle pairs via the \ifmcolor{superposition principle}~\eqref{eq:field_full},
whereas the CFM \cfmcolor{global drift}, defined through a \cfmcolor{conditional expectation}~\eqref{eq:cfm_ode_and_drift}, is generally
intractable to evaluate directly.
Second, the training procedures differ:
CFM uses a \cfmcolor{single-sample conditional velocity as the target}~\eqref{eq:cfm_objective} and learns it at points drawn from the \cfmcolor{conditional distribution},
while IFM uses the \ifmcolor{normalized global field as the target}~\eqref{eq:ifm_objective} and learns it at points drawn from a heuristic \ifmcolor{volume coverage distribution}.
Despite these differences, the following example (\S\ref{sec:pfgm}) suggests that the two viewpoints may describe the same underlying dynamics.

\subsection{Poisson Flow Generative Models}
\label{sec:pfgm}
PFGM is a noise-to-data generative framework rooted in electrostatics (akin to EFM).
It is of particular interest here because it admits both a field-based and a flow-based interpretation.
To our knowledge, PFGM is the only known framework that supports both viewpoints.

\ifmcolor{Field-based interpretation.}
PFGM embeds the data distribution into an extended space $\real^{\dim+1}$ by placing each sample $\x_0\sim\pi_0$ on the plane $\{z=0\}$, i.e., at $\xt_0=(\x_0,0)$.
Each $\xt_0$ is treated as a unit positive charge, inducing an electric field in accordance with Coulomb's law:
\begin{align}
    \E^{\x_0}(\xt) \propto {(\xt - \xt_0)}/{\|\xt - \xt_0\|_2^{\dim + 1}},
\end{align}
for $\xt=(\x,z)\in\real^{\dim+1}$.
The global field $\E$ is obtained via the \textit{electrostatic superposition principle}.
The field lines of $\E$ then define a transport from the data distribution to a prior $\pi_L$, which is a uniform distribution over a hemisphere of radius $L$ (with $L\to\infty$ or sufficiently large in practice).

\textit{Remark.}
Despite its resemblance to IFM, PFGM is not an instance of the two-sided
interaction field defined above: its conditional field is one-sided,
constructed from a single target point $\x_0$.
In~\S\ref{sec:one_sided_ifm} we introduce a one-sided notion of an
interaction field, of which PFGM is an instance.

\cfmcolor{Flow-based interpretation.}
Subsequent work~\citep[PFGM++]{pfgm++} reveals that PFGM possesses an underlying flow-based structure. Specifically, it demonstrates that PFGM's data-space generative dynamics, as in~\eqref{eq:ifm_sampling}, can be obtained by minimizing the objective
\begin{align}\label{eq:pfgmpp_objective}
    \mathbb{E}_{z,\x_0,\x_z}\,
    \left\|f_\theta(\x_z, z) - {(\x_z-\x_0)}/{z}\right\|_2^2,
    \quad
    p_z^{\x_0}(\x) \propto
        \left(
            \|\x-\x_0\|_2^2 + z^2
        \right)^{-\frac{\dim+1}{2}},
\end{align}
where $z \sim \mathcal{U}[0, L]$ and $\x_z \sim p_z^{\x_0}$.
The similarity between~\eqref{eq:pfgmpp_objective} and the \CFM{} objective~\eqref{eq:cfm_objective} suggests a flow-based interpretation of PFGM: $p_z^{\x_0}$ in~\eqref{eq:pfgmpp_objective} acts as a conditional path distribution, and $(\x_z-\x_0)/z$ serves as the corresponding conditional velocity target.
\begin{takeawaybox}
\new{\textbf{Summary.}}
The PFGM example shows that one generative transport can be described either
as a \emph{flow} or as \emph{field lines}. This motivates three questions:
\begin{enumerate}[leftmargin=*, align=left, label=\textbf{(\arabic*)}]
    \item
    PFGM is one-sided: can \IFM{} be given a one-sided formulation, as
    \CFM{} has? (\S\ref{sec:one_sided_ifm})

    \item
    Is this dual description specific to PFGM, or is there a
    \CFM{}--\IFM{} duality?
    (\S\ref{sec:from_flows_to_fields},~\S\ref{sec:from_fields_to_flows})

    \item
    Can general \IFM{} realize data-space transport beyond this duality?
    (\S\ref{sec:beyond_foifm})
\end{enumerate}
\end{takeawaybox}

\section{Flow--field duality: \CFM{} $\equiv$ forward-only \IFM{}}\label{sec:duality}
\new{In this section, we take up the first two questions raised in~\S\ref{sec:pfgm}
and formally establish the duality between \CFM{} and forward-only \IFM{}.}
We first introduce in~\S\ref{sec:one_sided_ifm} a one-sided formulation of
\IFM{}, paralleling the one-sided setting in \CFM{} and enabling a unified treatment
of both one-sided and two-sided constructions. We then prove the two directions of
the equivalence. In~\S\ref{sec:from_flows_to_fields}, we show how any \CFM{}
construction induces a forward-only interaction field with the same data-space
dynamics. Conversely, in~\S\ref{sec:from_fields_to_flows}, we show how any
forward-only interaction field can be normalized into a conditional probability
path and velocity field, thereby recovering a \CFM{} representation.
\new{Together, the two directions establish the equivalence; the third question,
what lies outside the forward-only class, is taken up
in~\S\ref{sec:beyond_foifm}.}
All \underline{proofs} are provided in Appendix~\ref{app:duality_proofs}.

Before stating the \new{duality}, we make a simple observation.
In forward-only fields, the spatial coordinate $z$ (the distance from the target plate) plays the same role as the time variable $t$ in conditional flows.
Likewise, the plate separation $L$ corresponds to the time horizon $T$.
With this correspondence in mind, we will use the $(t,T)$-based notation for
conditional flows and interaction fields alike throughout the remainder of the
paper.

\subsection{One-sided IFM}
\label{sec:one_sided_ifm}
Standard \IFM{} is defined in a two-sided form. To state the duality for both one-sided and two-sided constructions, we begin by introducing a one-sided notion of an interaction field.
Following PFGM (\S\ref{sec:pfgm}) and the general \IFM{} formulation (\S\ref{sec:interaction_field_matching}), we define the one-sided interaction field on the extended space $\real^{\dim+1}$, where the data distribution lies on a hyperplane $\{t = 0\}$.

\textbf{\new{Definition (One-sided Interaction Field)}.}
A vector field $\E^{\x_0}(\xt) \in \real^{\dim+1}$ is a \emph{one-sided Interaction Field} if it satisfies the following three properties.
First, the field has a fixed total flux~\eqref{eq:ifm_total_flux} through any
closed surface that encloses $\xt_0$ and lies in the region $\{t<T\}$.
Second, the field is divergence-free:
$\nabla \cdot \E^{\x_0}(\xt) = 0$ for all $\xt$ with $t<T$, $\xt \neq \xt_0$.
Third, all its field lines, except for a family carrying zero flux through every transversal hypersurface (\S\ref{sec:interaction_field_matching}), start at the particle $\xt_0 = (\x_0, 0)$:
$
{d\xt(\tau)}/{d\tau} = \E^{\x_0}(\xt(\tau)),
\lim_{\tau\downarrow \tau_s} \xt(\tau) = \xt_0$,
where $\tau_s$ denotes the starting value of the field-line parameter,
and reach the plate $\{t = T\}$:
there exists $\tau_f > \tau_s$ such that $\xt(\tau_f) \in \real^{\dim} \times \{T\}$.

\textit{Remark.}
This definition matches the two-sided Interaction Field, except that the field lines are not required to end at any predetermined particle.

\textbf{Generative dynamics.}
The global field induced by the distribution $\pi_0$ is obtained via generalized superposition, analogously to~\eqref{eq:field_full}:
$
\E(\xt)
= \int \E^{\x_0}(\xt)\, \pi_{0}(\x_0)\, d\x_0.
$
The corresponding field lines~\eqref{eq:ifm_field_lines} push forward $\pi_0$ to a distribution $\pi_T$.
Consequently, one-sided IFM defines a source (prior) distribution $\pi_T$ and transports between $\pi_T$ and $\pi_0$ along the global field lines.

\textit{Remark.}
In contrast to two-sided IFM, the generative dynamics in one-sided IFM is essentially built into the construction, since the source (prior) distribution $\pi_T$ is defined as the pushforward induced by the field lines.
In the two-sided setting, both endpoint distributions are specified in advance, and the fact that the field lines induce transport between them is a nontrivial property.

\textbf{Example (PFGM $\in$ forward-only one-sided IFM).}
The PFGM field defined in~\S\ref{sec:pfgm} is an instance of one-sided \IFM{}.
Moreover, it is forward-only, i.e., $\E^{\x_0}(\xt)_t>0$ for all $t>0$.

\textit{Notation.}
As in \CFM{}, we use \condcolor{blue} to indicate that the source-conditioning
point is omitted, corresponding to the one-sided case. Throughout the remainder
of this section, $\ccond$ denotes either $\x_0$ in the one-sided case or
$(\x_0,\x_T)$ in the two-sided case.

\subsection{From Flows to Fields}
\label{sec:from_flows_to_fields}
We first establish the flow-to-field direction of the duality.

\begin{theorem}\label{thm:cfm_to_ifm}
\textnormal{(CFM $\subseteq$ forward-only IFM).}
Let $\pi_0$ and $\pi_T$ be $\dim$-dimensional target and source distributions, respectively.
Let $(\vcond(\x),\pcond(\x))$ be a Conditional Flow (\S\ref{sec:conditional_flow_matching}),
and let $\v_t(\x)$ and $p_t(\x)$ denote the global velocity~\eqref{eq:cfm_ode_and_drift} and marginal probability path~\eqref{eq:cfm_marginals} obtained by marginalizing over $\cjoint(\ccond)$, respectively.
Define the $(\dim+1)$-dimensional vector field
\begin{align}\label{eq:cfm_to_ifm}
\E^\ccond(\x,t)
:=
(
    {\vcond(\x)\,\pcond(\x)}, \pcond(\x)
).
\end{align}
Then, under the assumptions specified in Appendix~\ref{assumptions_1}:

\begin{enumerate}[leftmargin=*, align=left]
    \item
    The vector field $\E^\ccond$ is a forward-only Interaction Field (\S\ref{sec:interaction_field_matching})
    with unit total flux $\Phi_0 = 1$.
    
    \item
    The global field $\E$ obtained by the generalized superposition principle~\eqref{eq:field_full} can be recovered as
    \begin{align}
        \E(\x,t) = \big(\v_t(\x)\,p_t(\x),\, p_t(\x)\big).
    \end{align}
    
    \item The \CFM{} generative dynamics~\eqref{eq:cfm_ode_and_drift} coincide with the dynamics induced by field lines~\eqref{eq:ifm_field_lines}, reparameterized by the distance from the plate~\eqref{eq:ifm_sampling}:
    \begin{align}
        {d\x}/{dt} = \v_t(\x) = {\E(\x,t)_\x}/{\E(\x,t)_t}.
    \end{align}
\end{enumerate}
\end{theorem}
Our Theorem~\ref{thm:cfm_to_ifm} shows that every conditional flow admits a
forward-only field representation with the same data-space generative dynamics.
In this representation, the CFM probability current becomes the spatial
component of the interaction field, and the marginal density becomes its time
component. Concrete field realizations of common flow-based approaches, including
stochastic interpolants, diffusion models, and flow matching, are summarized in
Appendix~\ref{app:dual_constructions}, Table~\ref{tab:cfm_ifm_dual_constructions}.

\subsection{From Fields to Flows}
\label{sec:from_fields_to_flows}
We now establish the converse field-to-flow direction of the duality.
Motivated by Theorem~\ref{thm:cfm_to_ifm}, a natural attempt to recover a conditional flow from a given forward-only interaction field $\E^\ccond(\x, t)$ would be to set
\begin{align}\label{eq:ifm_to_cfm_assumption}
    \vcond(\x) = {\E^\ccond(\x, t)_\x}/{\E^\ccond(\x, t)_t},
    \quad
    \pcond(\x) = \E^\ccond(\x, t)_t.
\end{align}
However, the last component $\E^\ccond(\x,t)_t$ of an interaction field is not
necessarily normalized:
\begin{align}\label{eq:flux_across_slice}
    \Phi_t^\ccond \coloneqq \int_{\real^\dim} \E^\ccond(\x,t)_t\,d\x
\end{align}
need not equal $1$, and may, a priori, depend on $\ccond$ and $t$.
Nevertheless, it is in fact independent of both, as follows from the proposition below.
\begin{proposition}\label{thm:flux_conservation_across_slices}
\textnormal{(Flux Conservation across Slices).}
Under the assumptions specified in Appendix~\ref{assumptions_2},
the scalar $\Phi_t^\ccond$ in~\eqref{eq:flux_across_slice} is the flux through the slice $\{t=\mathrm{const}\}$ at distance $t$ from the target plate and coincides with the total flux $\Phi_0$~\eqref{eq:ifm_total_flux} of the forward-only Interaction Field.
\end{proposition}
Since $\Phi_0$ is a positive constant, dividing by it rescales the field
without changing its lines, and hence the induced dynamics.
We now establish the converse field-to-flow direction of the duality.
\begin{theorem}\label{thm:ifm_to_cfm}
\textnormal{(forward-only IFM $\subseteq$ CFM).}
Let $\pi_0$ and $\pi_T$ be $\dim$-dimensional target and source distributions, respectively.
Let $\E^\ccond(\x, t)$ be a forward-only Interaction Field,
and let $\E(\x, t)$ denote the global field obtained by the generalized superposition principle~\eqref{eq:field_full}.
Define the conditional velocity and conditional density path by
\begin{align}\label{eq:ifm_to_cfm}
    \vcond(\x) = {\E^\ccond(\x, t)_\x}/{\E^\ccond(\x, t)_t},
    \quad
    \pcond(\x) = {\E^\ccond(\x, t)_t}/{\Phi_0}.
\end{align}

Then, under the assumptions specified in Appendix~\ref{assumptions_2}:
\begin{enumerate}[leftmargin=*, align=left]
    \item
    The pair $(\vcond\!,\pcond)$ is a Conditional Flow (\S\ref{sec:conditional_flow_matching}).
    
    \item
    The global drift $\v_t$~\eqref{eq:cfm_ode_and_drift} and probability path $p_t$~\eqref{eq:cfm_marginals} can be recovered from the global field $\E$ via
    \begin{align}\label{eq:ifm_to_cfm_global}
        \v_t(\x) = {\E(\x,t)_\x}/{\E(\x,t)_t},
        \quad
        p_t(\x) = {\E(\x,t)_t}/{\Phi_0}.
    \end{align}
    
    \item
    Consequently, as in Theorem~\ref{thm:cfm_to_ifm}, the \CFM{} generative dynamics~\eqref{eq:cfm_ode_and_drift} coincide with the dynamics induced by field lines~\eqref{eq:ifm_field_lines}, reparameterized by the distance from the plate~\eqref{eq:ifm_sampling}.
\end{enumerate}

\end{theorem}
Theorem~\ref{thm:ifm_to_cfm} shows that every forward-only interaction field
admits a conditional flow representation with the
same data-space dynamics. The key step is the normalization of the
field's time component by the conserved flux $\Phi_0$, which turns the slice flux
density into a conditional probability path.

\textbf{Example (PFGM's flow-based construction).}
Previously, we noted that PFGM is a forward-only one-sided \IFM{} instance.
Therefore, by Theorem~\ref{thm:ifm_to_cfm}, it admits an equivalent conditional
flow representation. This construction recovers the objects introduced
in~\S\ref{sec:pfgm} as the flow-based interpretation of PFGM. In this sense, the
flow-based formulation used in PFGM++ arises naturally as a special case of the
field-to-flow construction.

\textbf{Example (IFM's flow-based construction).}
In~\S\ref{sec_extra:original_ifm_realization}, the canonical IFM field is constructed directly as a forward-only interaction field.
Using our Theorem~\ref{thm:ifm_to_cfm}, we can express it as a conditional flow, which has a simpler form:
$\v_t^\cond(\x_t) = \Idot{t} + ({\sdot{t}}/{\s{t}}) \big(\x_t - \I{t}\big)$,
$p_t^\cond(\x) = \mathcal{N}(\x \mid \I{t}, \s{t}^2 \Id)$,
where $\I{t}$ and $\s{t}$ are specified in Appendix~\ref{app:dual_constructions}, Table~\ref{tab:cfm_ifm_dual_constructions}.
This representation makes explicit that the IFM field is induced by the two-sided stochastic interpolant from~\S\ref{sec_extra:two_sided_stochastic_interpolant}, namely, $\x_t = \I{t} + \s{t}\e$.
Additional field--flow construction examples are summarized in the same table.
\section{Beyond forward-only IFM}
\label{sec:beyond_foifm}

The constructions in \S\ref{sec:duality} establish an equivalence between
CFM and forward-only IFM at the conditional level. Since they commute
with marginalization and superposition, this equivalence extends to the
induced global data-space dynamics. This leaves open whether general IFM can realize data-space transport
beyond this equivalence. Backward-oriented field lines alone do not
establish such a separation: different geometries in the augmented space
may still induce the same data-space transport.
All \underline{proofs} are provided in Appendix~\ref{app:separation_proofs}.

To prove a strict separation, we therefore identify an obstruction at the
level of data-space transport maps. We approach this question through
orientation: a differentiable map $G:\real^\dim\to\real^\dim$ preserves
(reverses) orientation on an open set if $\det\nabla G>0$ ($<0$) there.
In one dimension, this means preserving (reversing) the order of nearby
points; for example, $G(a)=-a$ is orientation-reversing.

Let $G_{s,t}$ denote the flow map of the global velocity $\v_t$ from time
$s$ to $t$, i.e., $\partial_t G_{s,t}(a)=\v_t(G_{s,t}(a))$ with
$G_{s,s}(a)=a$.
Since $\v_t$ may be singular at an endpoint carrying a compactly supported
density~\citep[\S3.5]{lipman2024flow}, 
we only assume it to be smooth on
$(0,T)$ and take the CFM transport map $G_{T,0}$ to be the locally uniform
limit of $G_{s,t}$ as $s\uparrow T$, $t\downarrow 0$, wherever it exists.
Each $G_{s,t}$ with $0<t<s<T$ is smooth and
injective~\citep[Thm.~1]{lipman2024flow}, with
$\det\nabla G_{s,t}(a)=\exp\!\big(\int_s^t(\nabla\!\cdot\v_\tau)(G_{s,\tau}(a))\,d\tau\big)>0$
by Liouville's formula.
A locally uniform limit of such orientation-preserving maps
cannot reverse orientation: $\det\nabla G_{T,0}\ge 0$ wherever $G_{T,0}$ is
continuously differentiable (a standard consequence of degree
theory~\citep[Def.~2.1 and Thm.~3.1]{deimling1985nonlinear}).

General IFM admits a different behavior.
A generative trajectory starting at $t=T$ may reach the target
plate $t=0$, continue into $t<0$, and then return to $t=0$.
As we show below, the resulting transport map reverses
orientation.
\begin{proposition}[Returning trajectories reverse orientation]
    \label{prop:ifm_return_orientation}
    Let $\E$ be a global IFM field and let $U\subset\real^\dim$ be a nonempty
    open set. Suppose the generative trajectories issued from $U\times\{T\}$
    enter the region between the plates, cross the target plate, and return
    to it from $t<0$, and let $F:U\to\real^\dim$ be the resulting transport
    map. Then, under
    regularity and transversality assumptions specified in
    Appendix~\ref{app:return_orientation}, $F$ reverses orientation.
\end{proposition}
The next proposition shows that EFM realizes this mechanism: its sampling
rule selects such a returning trajectory with positive probability.
\begin{proposition}[EFM admits orientation-reversing trajectories]
\label{prop:efm_return_example}
For every $D\ge1$, there exist smooth, compactly supported source and
target probability densities and a nonempty open set
$U\subset\real^\dim$ with $\pi_T(U)>0$ such that, under the IFM sampling
rule~\citep[Appendix~A.2]{manukhov2025interaction} in the conventions of
Appendix~\ref{app:efm_assumptions}, a trajectory initialized
at any $a\in U$ crosses the target plate, enters $t<0$, returns, and
terminates at its second crossing with probability one. The resulting
transport map $F:U\to\real^\dim$ reverses orientation:
$\det\nabla F(a)<0$ for every $a\in U$.
\end{proposition}
Together with the orientation property of CFM, Proposition~\ref{prop:efm_return_example}
yields the desired separation.
\begin{corollary}[Strict separation]
\label{cor:separation}
The EFM transport map $F$ of
Proposition~\ref{prop:efm_return_example} is realized by no CFM
flow, for any source and target densities: $G_{T,0}$ and $F$ do not
coincide on any nonempty open subset of $U$.
Combined with \S\ref{sec:duality}, general IFM is therefore strictly more
expressive than CFM at the level of data-space transport maps.
\end{corollary}
Indeed, if $G_{T,0}=F$ on a nonempty open $V\subset U$, then $G_{T,0}$ is
continuously differentiable on $V$ with
$\det\nabla G_{T,0}=\det\nabla F<0$, contradicting the orientation
property of CFM.
\def\sigmaval{0.05}
\section{\new{Duality Takeaways}}
\label{sec:duality_takeaways}
We now collect the main implications of the flow--field duality.
The section is organized around three takeaways:
for each one we state the implication and briefly discuss why it matters.
Each is supported empirically on a controlled setting, described together with the results in
Appendix~\ref{app:experiments}.
Additional discussions are provided in Appendix~\ref{app:additional_discussion_and_experiments}.

\begin{takeawaybox}
    \textbf{Takeaway 1 (Expressiveness).}
    At the level of induced data-space transport maps,
    \[
        \mathrm{CFM}
        \equiv
        \mathrm{forward\text{-}only\ IFM}
        \subsetneq
        \mathrm{IFM}.
    \]
\end{takeawaybox}

Theorems~\ref{thm:cfm_to_ifm} and~\ref{thm:ifm_to_cfm} show that CFM
and forward-only IFM induce the same class of data-space
generative dynamics. Thus, the probabilistic CFM view and the geometric
IFM view are not separate model classes in this regime: they are two
representations of the same dynamics.

The strict inclusion comes from trajectories that cross the target plate and
return: Proposition~\ref{prop:ifm_return_orientation} shows that their transport
maps reverse orientation, whereas no CFM transport map does.
This already occurs in EFM, with positive
probability (Proposition~\ref{prop:efm_return_example}).
This is the concrete property that puts general IFM outside CFM, and the reason
it is worth developing on its own.

Appendix~\ref{app:duality_verification} verifies the equivalence
of~\S\ref{sec:duality} directly: we train a CFM drift and its dual interaction
field independently, and compare the data-space dynamics they induce. Built from
the same conditional flow, the two agree to a mean squared error of order
$10^{-7}$; dynamics built from different conditional flows differ by
$4.9\times10^{-1}$, six orders of magnitude more.
The equivalence therefore survives estimation: it is not only an identity between
exact objects, but something two separately trained models arrive at.

\begin{takeawaybox}
    \textbf{Takeaway 2 (IFM probabilistic interpretation).}
    A global \emph{forward-only} interaction field $\E(\x,t)$ induces not only
    field-line dynamics, but also an explicit data-space distribution path:
    \[
        p_t(\x) = \E(\x,t)_t / \Phi_0 .
    \]
\end{takeawaybox}
Theorem~\ref{thm:ifm_to_cfm} shows that the geometric IFM representation has an
equivalent probabilistic view.
Namely, for the forward-only fields the information about the induced probabilistic path is carried by the last component of the global field.

Appendix~\ref{app:probabilistic_interpretation} illustrates this
correspondence empirically. We compare particle distributions
generated by the trained IFM model with heatmaps of the last
component of the global field, estimated by Monte Carlo averaging
over sampled endpoint pairs.
Their qualitative agreement
(Figure~\ref{fig:ifm_probabilistic_interpretation})
supports the probabilistic interpretation predicted by the
flow--field duality.

\begin{takeawaybox}
    \textbf{Takeaway 3 (Field-informed volume coverage).}
    \emph{Forward-only} IFM induces a natural sampling distribution over the augmented
    space: sample $t \sim \mathcal{U}[0,T]$ and then
    \[
        \x \sim p_t^\ccond(\x),
        \qquad
        p_t^\ccond(\x)
        =
        \E^\ccond(\x,t)_t/\Phi_0 .
    \]
\end{takeawaybox}
Training an interaction field requires a distribution $p_{\mathrm{vol}}$ covering
the region where the field is defined~\eqref{eq:ifm_objective}, and IFM offers no
principled choice: it is picked heuristically, one of the differences from CFM
noted in~\S\ref{sec:cfm_vs_ifm}.

Substituting the duality~\eqref{eq:ifm_to_cfm} into the CFM
objective~\eqref{eq:cfm_objective} gives
\begin{align}\label{eq:ifm_single_sample_objective}
    \mathbb{E}_{t,\ccond,\,\x \sim p_t^\ccond}
    \left\|
        \v_\theta(\x,t)
        -
        {\E^\ccond(\x,t)_\x}/{\E^\ccond(\x,t)_t}
    \right\|_2^2 .
\end{align}
This has two consequences. First, the dynamics induced by a forward-only interaction field
can be learned by single-sample regression: the target
in~\eqref{eq:ifm_single_sample_objective} is built from one endpoint pair, whereas
the field target in~\eqref{eq:ifm_objective} has to be estimated from a batch.
Second, the objective arrives with its own sampling distribution
$p_t^\ccond = \E^\ccond(\x,t)_t/\Phi_0$, which is exactly the principled
$p_{\mathrm{vol}}$ that IFM was missing. It can be used inside IFM as well,
replacing the heuristic volume distribution in the normalized-field objective:
\begin{align}\label{eq:ifm_objective_with_new_volume}
    \mathbb{E}_{t,\ccond,\,\x \sim p_t^\ccond}
    \left\|
        f_\theta(\x,t)
        -
        {\E(\x,t)}/{\|\E(\x,t)\|_2}
    \right\|_2^2 .
\end{align}
Appendix~\ref{app:volume_coverage_ablation} checks the second consequence: we fix
the normalized-field objective~\eqref{eq:ifm_objective} and the target field, and
vary only $p_{\mathrm{vol}}$, comparing the field's own conditional path, which
turns the objective into~\eqref{eq:ifm_objective_with_new_volume}, with the path
induced by a different field.
Its own path more than halves the final $W_2$ for
both fields we test ($0.58 \to 0.25$ and $0.81 \to 0.38$).
The choice of $p_{\mathrm{vol}}$ is therefore not a harmless implementation
detail: it costs accuracy when it does not match the target field, and the duality supplies the matching one without an additional
heuristic choice.

\textbf{Conclusion.}
In this work we reveal the relation between two frameworks, CFM and IFM, that
are built on different principles.
We show that on the forward-only subclass they are one theory, constructively
and in both directions.
This duality removes the reason to study the two frameworks apart in that regime,
and its field-to-flow direction gives IFM two things it lacked: an explicit
marginal path and a training distribution read off the field itself.
The inclusion, however, is strict: general IFM is not a reformulation of CFM but
a genuinely larger class, so models outside the forward-only subclass can do
things no conditional flow can, which makes general IFM a direction worth
developing on its own.
The separating property is concrete, orientation reversal, and EFM already
realizes it, which completes the taxonomy in Figure~\ref{fig:teaser}.

% \section*{Limitations}
% Our main contribution is theoretical. The empirical results are intended to validate the duality and illustrate estimator behavior, not to establish a new state-of-the-art generative model.

\section*{Acknowledgments}
The work was supported by the grant for research centers in the field of AI provided by the Ministry of Economic Development of the Russian Federation in accordance with the agreement 000000C313925P4F0002 and the agreement No.~139-10-2025-033.

\clearpage
\bibliographystyle{unsrtnat}
\bibliography{bibliography}

%%%%%%%%%%%%%%%%%%%%%%%%%%%%%%%%%%%%%%%%%%%%%%%%%%%%%%%%%%%%
\newpage
\appendix
% !TeX root = ../main.tex
\section{Additional material for Section~\ref{sec:background}}
\label{app:preliminaries}

\subsection{CFM vs.\ IFM: side-by-side comparison}
\label{app:cfm_vs_ifm}
Table~\ref{tab:cfm_vs_ifm} lists the main terms of the two frameworks side by side,
as discussed in~\S\ref{sec:cfm_vs_ifm}.

\newcolumntype{A}{>{\raggedright\arraybackslash\columncolor{generalBG}}p{2.05cm}}
\newcolumntype{C}{>{\centering\arraybackslash\columncolor{cfmBG}}X}
\newcolumntype{I}{>{\centering\arraybackslash\columncolor{ifmBG}}X}

\begin{table*}[!ht]
\centering
% \caption{\textbf{CFM vs.\ IFM:} key conceptual differences. Notation and equation references follow the Background section.}
\caption{\textbf{CFM vs.\ IFM}: side-by-side comparison of the Conditional Flow Matching and Interaction Field Matching frameworks.}
\fontsize{6.0}{9}\selectfont   % <--- custom size
\setlength{\tabcolsep}{6pt}    % optional: tighter columns
\renewcommand{\arraystretch}{1.25}
\setlength{\abovedisplayskip}{4pt}
\setlength{\belowdisplayskip}{4pt}
\begin{tabularx}{\textwidth}{A C I}
\rowcolor{generalBG}
\textbf{Aspect} &
\textbf{\textcolor{cfmcolor}{Conditional Flow Matching}} &
\textbf{\textcolor{ifmcolor}{Interaction Field Matching}} \\
\midrule

\textbf{Elementary object}
&
% \textbf{Conditional Flow} (Eqs.~\eqref{eq:cfm_cond_transport_equation}--\eqref{eq:cfm_cond_boundaries})
\textbf{Conditional Flow} (\S\ref{sec:conditional_flow_matching})
\[
(\vcond(\x),\, \pcond(\x))
% (\v_t^{\cond}(\x),\, p_t^{\cond}(\x))
\]
defined \textbf{in the data space} \(\real^\dim\)
&
% \textbf{Interaction Field} (Eqs.~\eqref{eq:if_connects_particles}--\eqref{eq:if_conserves_flux})
\textbf{Interaction Field}~\eqref{eq:if_connects_particles}
\[
\E^{\condl}(\xt)
\]
defined \textbf{in the extended space} \(\real^{\dim+1}\)
\\ \addlinespace[2pt]
\textbf{Global object}
&
{\setlength{\belowdisplayskip}{0pt}
\textbf{Conditional averaging} of conditional velocities:
$\displaystyle
\begin{aligned}
    \! \! \! \!
    \v_t(\x) 
    &= \! \int\vcond(\x) p_t(\ccond \! \mid \! \x_t \! = \!\x)d\x_0d\condcolor{\x_T}
    \\
    &\neq \! \int\vcond(\x)\,\pi_{0,T}(\ccond)\,d\x_0\,\condcolor{d\x_T}
\end{aligned}
$
}
% \vspace{-3pt}
&
{\setlength{\belowdisplayskip}{0pt}
\textbf{Generalized superposition} of pairwise fields:
\[
\E(\xt)=\int \E^{\condl}(\xt)\,\pi_{0,L}(\condl)\,d\x_0\,d\x_L
\]
}
% \vspace{-3pt}
\\ \addlinespace[2pt]
\textbf{Generative dynamics}
&
ODE-based dynamics \textbf{in the data space}:
\[
\frac{d\x}{dt}=\v_t(\x), \quad \x_T \sim \pi_T
\]
&
Field-line transport \textbf{in the extended space}:
\[
\frac{d\xt}{d\tau}=\E(\xt), \quad \xt \sim \pi_L \times \delta_{z=L}
\]
{\setlength{\belowdisplayskip}{0pt}
\textbf{Forward-only} fields dynamics \textbf{in the data space}:
\[
\frac{d\x}{dz}=\frac{\E(\xt)_\x}{\E(\xt)_z}, \quad \x_L \sim \pi_L
\]}
\\ \addlinespace[2pt]
\textbf{Marginal distributions}
&
{\setlength{\belowdisplayskip}{0pt}
\textbf{Explicitly} defined via conditional distributions:
\[
p_t(\x)=\int \pcond(\x)\,\pi_{0,T}(\ccond)\,d\x_0\,\condcolor{d\x_T}
\]
}
% \vspace{-3pt}
&
{\setlength{\belowdisplayskip}{0pt}
\textbf{No explicit} marginal density path:
\[
\phantom{\int}\text{defined implicitly via field-line transport~\eqref{eq:ifm_field_lines}}\phantom{\int}
\]
}
% \vspace{-3pt}
\\ \addlinespace[2pt]
\textbf{Training objective}
&
{\setlength{\belowdisplayskip}{0pt}
\textbf{Single-sample conditional velocity} regression:
\[
\mathbb{E}_{t,\,\ccond,\,\x_t}
\left\|
\v_t(\x_t)-\vcond(\x_t)
\right\|_2^2
\]
}
% \vspace{-3pt}
&
\textbf{Multi-sample global field} regression:
\[
\mathbb{E}_{\xt \sim p_{\mathrm{vol}}}
\left\|
f_\theta(\xt)-\E(\xt)
\right\|_2^2
\]
{\setlength{\belowdisplayskip}{0pt}
\textbf{Normalized field} in the forward-only case:
\[
\mathbb{E}_{\xt \sim p_{\mathrm{vol}}}
\left\|
f_\theta(\xt)-\frac{\E(\xt)}{\|\E(\xt)\|_2}
\right\|_2^2
\]
}
% \vspace{-3pt}
\\ \addlinespace[2pt]

\textbf{Training volume}
&
{\setlength{\belowdisplayskip}{-6pt}
\textbf{Explicitly} defined by conditional distributions:
\[
\ccond \sim \cjoint,\
t\sim \mathcal{U}[0,T],\
\x_t \sim \pcond
\]
}
% % \vspace{-3pt}
&
{\setlength{\belowdisplayskip}{-6pt}
\textbf{Ad hoc heuristic} distribution:
\[
\xt \sim p_{\mathrm{vol}}
\]
}
% % \vspace{-3pt}
\\
\bottomrule
\end{tabularx}
\label{tab:cfm_vs_ifm}
\vspace{-10pt}
\end{table*}

\subsection{Original IFM field realization}
\label{app:original_ifm_field}
For completeness, we restate the forward-only Interaction Field realization proposed in
\citet[Appendix A.4]{manukhov2025interaction}.
They construct $\Econd$ in two steps. For a pair on a common axis,
$\xt_0=(\mathbf{0},0)$ and $\xt_L=(\mathbf{0},L)$, split an arbitrary point as
$\xt = \xt_\perp + z\,\mathbf{e}_z$, where $\mathbf{e}_z$ is the unit vector
along the $z$-axis and $\xt_\perp$ is the component of $\xt$ orthogonal to it,
and write $x_\perp := \|\xt_\perp\|_2$ and
$\mathbf{e}_\perp := \xt_\perp / x_\perp$.
The field is then given by a unit direction $\n^\condl(\xt)$ and a magnitude
$\|\Econd(\xt)\|_2$:
\begin{gather}\label{eq:original_ifm_field}
    \n^\condl(\xt) := \cos\alpha(\xt)\,\mathbf{e}_z
        + \sin\alpha(\xt)\,\mathbf{e}_\perp,
    \\[2pt]
    \|\Econd(\xt)\|_2 :=
    \frac{\exp\!\left(-\frac{x_\perp^2}{2\sigma(z)^2}\right)}{\sigma(z)^\dim \cos\alpha(\xt)},
    \\[2pt]
    \sigma(z) := \sin\!\left(\frac{\pi z}{L}\right),
    \qquad
    \alpha(\xt) := \arctan\!\left(
        \frac{\pi}{L}\,x_\perp \cot\!\left(\frac{\pi z}{L}\right)
    \right),
\end{gather}
where $\sigma(z)$ is the width of the tube at height $z$
(for $d = L/2$ and $\sigma_0 = 1$ in their notation)
and the angle $\alpha(\xt)$, measured from the $z$-axis, is chosen so that the
ratio $x_\perp / \sigma(z)$ remains constant along each field line.
For a pair in general position, $\xt_0=(\x_0,0)$ and $\xt_L=(\x_L,L)$, the field
is obtained from the above by a shift parallel to the plates, carrying the axis
onto the segment joining $\xt_0$ and $\xt_L$ and leaving the height $z$
unchanged.

% \newpage

\section{Additional material for Section~\ref{sec:duality}}
\label{app:duality_proofs}
This section collects the dual constructions referenced in~\S\ref{sec:duality}
together with the assumptions and proofs for its results.

\subsection{Flow--field dual constructions}
\label{app:dual_constructions}
In this section, we provide additional examples of common frameworks formulated via the flow-field dual construction, as summarized in Table~\ref{tab:cfm_ifm_dual_constructions}.

% Requires: \usepackage{array,tabularx,colortbl,makecell}

% X becomes vertically centered AND horizontally centered
\renewcommand\tabularxcolumn[1]{>{\centering\arraybackslash}m{#1}}

% Column types (colors kept; centering for X is now automatic via tabularxcolumn)
\newcolumntype{E}{>{\columncolor{generalBG}\centering\arraybackslash}m{2.2cm}}
\newcolumntype{F}{>{\columncolor{cfmBG}\centering\arraybackslash}X}
\newcolumntype{G}{>{\columncolor{ifmBG}\centering\arraybackslash}X}

% Optional: a compact header style helper
\newcommand{\colhead}[3][0pt]{%
  \makecell{\textbf{#2}\\[\dimexpr#1\relax]\scriptsize #3}%
}

\begin{table}[htb]
\centering
\caption{\textbf{Flow-field dual constructions}.
This table shows how a common framework can be formulated using both CFM and IFM.
}
\label{tab:cfm_ifm_dual_constructions}
\scriptsize
% \footnotesize
\setlength{\tabcolsep}{6pt}
\renewcommand{\arraystretch}{1.25}

\begin{tabularx}{\textwidth}{E F G}
\rowcolor{generalBG}
\colhead{Framework}{}
&
\colhead[3pt]{\textcolor{cfmcolor}{CFM structure}}{\big($\vcond(\x),\,\pcond(\x)\big)$}

&
\colhead[3pt]{\textcolor{ifmcolor}{IFM structure}}{$\!\!\Eccondt(\xt)\!=\!(\E^\ccond(\xt)_\x,\E^\ccond(\xt)_t)$,\\$\xt\!=\!(\x, t)$}
\\
\midrule
\makecell{\textbf{Flow Matching}\\\citep{lipman2022flow}}
&
$\displaystyle
\begin{gathered}
  \v_t^{\x_0}(\x) = \frac{\x-\x_0}{t}\\
  p_t^{\x_0}(\x) = \mathcal{N}\!\bigl(\x \mid (1-t)\x_0,\, t^2\Id\bigr)
\end{gathered}
$
&
$\displaystyle
\begin{gathered}
  \E^{\x_0}(\x,t)_\x
  =
  \v_t^{\x_0}(\x)\, p_t^{\x_0}(\x)
  \\[3pt]
  \E^{\x_0}(\x,t)_t
  =
  p_t^{\x_0}(\x)
\end{gathered}
$
\\ \addlinespace[2pt]
\makecell{\textbf{VE-diffusion / EDM}\\\citep{song2020score,karras2022elucidating}}
&
$\displaystyle
\begin{gathered}
  \v_t^{\x_0}(\x) = \frac{\sdot{t}}{\s{t}}\left(\x - \x_0\right)\\
  p_t^{\x_0}(\x) = \mathcal{N}\!\bigl(\x \mid \x_0,\, \s{t}^2\Id\bigr)
\end{gathered}
$
&
$\displaystyle
\begin{gathered}
  \E^{\x_0}(\x,t)_\x
  =
  \v_t^{\x_0}(\x)\, p_t^{\x_0}(\x)
  \\[3pt]
  \E^{\x_0}(\x,t)_t
  =
  p_t^{\x_0}(\x)
\end{gathered}
$
\\ \addlinespace[2pt]
\makecell{\textbf{Linear Interpolant}\\\citep{albergo2023stochastic}}
&
$\displaystyle
\begin{gathered}
  \!\!\!\v_t^{\x_0,\x_T}(\x)
    \!=\!
    \frac{\sdot{t}}{\s{t}} \!\left(
        \x - \mathcal{I}^\cond(t)
    \right) \!+\! \mathcal{\dot{I}}^\cond(t)\\ 
  p_t^{\x_0, \x_T}(\x) = \mathcal{N}(x \mid \I{t}, \s{t}^2 \Id)\\[10pt]
  \mathcal{I}^\cond(t) = \x_0 \cdot \left(1 - {t}/{T}\right) + \x_T \cdot {t}/{T}\\
  \s{t} = \sqrt{2\cdot{t}/{T}\cdot\left(1 - {t}/{T}\right)}
\end{gathered}
$
&
$\displaystyle
\begin{gathered}
  \E^{\x_0,\x_T}(\x,t)_\x
  =
  \v_t^{\x_0,\x_T}(\x)\, p_t^{\x_0,\x_T}(\x)
  \\[3pt]
  \E^{\x_0,\x_T}(\x,t)_t
  =
  p_t^{\x_0,\x_T}(\x)
\end{gathered}
$
\\ \addlinespace[2pt]
\makecell{\textbf{PFGM}\\\citep{pfgm}}
&
$\displaystyle
\begin{gathered}
  \v_t^{\x_0}(\x) = \frac{\x-\x_0}{t}\\
  p_t^{\x_0}(\x) \propto {t}/{
    \left(
        \|\x - \x_0\|_2^2 + t^2
    \right)^{\frac{D + 1}{2}}
}
\end{gathered}
$
&
$\displaystyle
\begin{gathered}
    \E^{\x_0}(\xt)=
    \frac{\xt-\xt_0}{\|\xt-\xt_0\|_2^{\dim+1}}
\end{gathered}
$
\\ \addlinespace[2pt]
\makecell{\textbf{PFGM++} ($\extradim \ge 1$)\\\citep{pfgm++}}
&
$\displaystyle
\begin{gathered}
  \v_t^{\x_0}(\x) = \frac{\x-\x_0}{t}\\
  p_t^{\x_0}(\x) \propto {t^{\extradim}}/{
    \left(
        \|\x - \x_0\|_2^2 + t^2
    \right)^{\frac{D + \extradim}{2}}
}
\end{gathered}
$
&
$\displaystyle
\begin{gathered}
  \E^{\x_0}(\x,t)_\x
  =
  \v_t^{\x_0}(\x)\, p_t^{\x_0}(\x)
  \\[3pt]
  \E^{\x_0}(\x,t)_t
  =
  p_t^{\x_0}(\x)
\end{gathered}
$
\\ \addlinespace[2pt]
\makecell{\textbf{IFM}\\\citep{manukhov2025interaction}}
&
$\displaystyle
\begin{gathered}
  \!\!\!\v_t^{\x_0,\x_T}(\x)
    \!=\!
    \frac{\sdot{t}}{\s{t}} \!\left(
        \x - \mathcal{I}^\cond(t)
    \right) \!+\! \mathcal{\dot{I}}^\cond(t)\\ 
  p_t^{\x_0, \x_T}(\x) = \mathcal{N}(x \mid \I{t}, \s{t}^2 \Id)\\[10pt]
  \mathcal{{I}}^\cond(t) = \x_0 \cdot \left(1 - {t}/{T}\right) + \x_T \cdot {t}/{T}\\
  \s{t} = \sin\left({\pi t}/{T}\right)
\end{gathered}
$
&
\makecell[c]{%
$\displaystyle
\begin{gathered}
  \E^{\x_0,\x_T}(\x,t)_\x
  =
  \v_t^{\x_0,\x_T}(\x)\, p_t^{\x_0,\x_T}(\x)
  \\[3pt]
  \E^{\x_0,\x_T}(\x,t)_t
  =
  p_t^{\x_0,\x_T}(\x)
\end{gathered}
$\\[15pt]
\scriptsize OR originally defined as in~\S\ref{sec_extra:original_ifm_realization}%
}
\\ \addlinespace[2pt]
\makecell{\textbf{EFM}\\\citep{kolesov2025field}}
&
{\Large$\times$}
&
$\displaystyle
\begin{gathered}
    \E^{\x_0,\x_T}(\xt) \!=\!
    \frac{\xt-\xt_0}{\|\xt-\xt_0\|_2^{\dim+1}}
    \!-\!
    \frac{\xt-\xt_T}{\|\xt-\xt_T\|_2^{\dim+1}}
\end{gathered}
$
\\
\bottomrule
\end{tabularx}
\end{table}
% \textbf{Example (PFGM $\in$ foward-only one-sided IFM).}
% \textbf{Example (PFGM's flow-based construction).}
% \textbf{Example (IFM's flow-based construction).}

\subsection{Assumptions for Theorem~\ref{thm:cfm_to_ifm}}
\label{assumptions_1}

For all $t \in (0,T)$, $\mathbf{x} \in \mathbb{R}^D$:

\begin{enumerate}
    \item \new{$\E^\ccond = (\mathbf{v}_t^{\ccond} p_t^{\ccond},\, p_t^{\ccond}) \in C^1((0,T) \times \mathbb{R}^D)$,
      and $p_t^{\ccond}(\mathbf{x})$ is strictly positive;}

    \item All integrals appearing in the proof (those involving $p_t^{\ccond}$ and $\mathbf{v}_t^{\ccond} p_t^{\ccond}$)
      converge; \new{in particular
      $\int_0^T\!\!\int_{\mathbb{R}^D}\|\mathbf{v}_t^{\ccond}(\mathbf{x})\|\,p_t^{\ccond}(\mathbf{x})\,d\mathbf{x}\,dt<\infty$,
      which is the integrability hypothesis of the superposition principle.}

\end{enumerate}

\neww{\textit{Boundary convention.}
In both directions of the duality, pointwise identities are understood on
the open slab $\real^\dim\times(0,T)$. After extension by zero, boundary
charges and fluxes are understood in the distributional and weak normal-trace
senses, respectively; no smooth extension across the plates is required.}

% \textit{Proof.}
\subsection{Proof of Theorem~\ref{thm:cfm_to_ifm}}
\label{proof_1}

We prove the three items in order.

\begin{enumerate}[itemsep=-1pt, topsep=-1pt, label=\textbf{\arabic*.}]

% ------------------------------------------------------------
\item \textbf{$\Eccondt(\x, t)$ is a forward-only Interaction Field with $\Phi_0=1$.}

\textit{Forward-only.}
\new{By definition of the field, $\Eccondt(\x,t)_t = p_t^\ccond(\x)$, which is strictly positive on
$(0,T)\times\real^\dim$ by Assumption~1. Outside the slab the field is extended by zero, so
$\Eccondt(\x,t)_t>0$ at every point where $\Eccondt$ is nonzero, and $\Eccondt$ is a forward-only
field.}

\textit{Field lines connect the two particles.}
\new{The connecting property~\eqref{eq:if_connects_particles} follows from the superposition principle
together with the boundary conditions~\eqref{eq:cfm_cond_boundaries}.}

\new{By the superposition principle for the continuity equation~\citep{bogachev2021ambrosio}, applied to the
pair $(\vcond,\pcond)$ solving~\eqref{eq:cfm_cond_transport_equation}, there exists a probability measure
$\eta$ on the space of continuous curves $C([0,T];\real^\dim)$, concentrated on absolutely continuous
curves $\x(\cdot)$ that solve
\[
\frac{d\x}{dt}=\vcond(\x(t))
\qquad\text{for a.e. } t\in(0,T),
\]
and such that $(e_t)_\#\eta=\pcond$ for every $t\in[0,T]$, where $e_t\big(\x(\cdot)\big):=\x(t)$ denotes
the evaluation map.}

\new{The boundary conditions~\eqref{eq:cfm_cond_boundaries} read
$(e_0)_\#\eta=\delta_{\x_0}$ and $\condcolor{(e_T)_\#\eta=\delta_{\x_T}}$.
A push-forward equal to a Dirac mass forces the corresponding evaluation to be $\eta$-almost surely
constant. \neww{Let $C$ be the set of absolutely continuous solutions
of the above ODE with $\x(0)=\x_0$ and
$\condcolor{\x(T)=\x_T}$. Then $\eta(C)=1$.}
Thus $\eta$-almost every curve of the representation runs from $\x_0$ to $\condcolor{\x_T}$; in
particular, at least one such curve exists. Such a curve need not be unique, and
Definition~\eqref{eq:if_connects_particles} does not require uniqueness: it only asks the field lines to
\emph{connect} the two particles, up to a family carrying zero flux. In typical constructions the
connecting family is in fact infinite (Remark~\ref{rem:many_characteristics}).}

\new{Fix any curve $\x(\cdot)\in C$} and put
\(
\xt(t):=(\x(t),t).
\)
It connects $\xt_0=(\x_0,0)$ to $\condcolor{\xt_T=(\x_T,T)}$ in $\real^{\dim+1}$ and follows
\[
\frac{d\xt}{dt}=(\vcond(\x(t)),1).
\]

Now note that
\[
\Eccondt(\x,t)=\big(\vcond(\x)\,p_t^\ccond(\x),\;p_t^\ccond(\x)\big)
= p_t^\ccond(\x)\,(\vcond(\x),1).
\]
Hence $\Eccondt(\xt)$ is a {positive scalar multiple} of $(\vcond(\x), 1)$ on $p_t^\ccond(\x)>0$, so it has the same integral curves
up to a reparameterization. Concretely, define a new parameter $\tau$ along the above curve by
\begin{align*}
\frac{dt}{d\tau}&=p_t^\ccond(\x(t))\qquad\Big(\text{equivalently }\,\frac{d\tau}{dt}
=\frac{1}{p_t^\ccond(\x(t))}\Big),
\end{align*}
and set $\xt(\tau):=\xt(t(\tau))$. Then
\begin{align*}
\frac{d\xt}{d\tau}
&=
\frac{d\xt}{dt}\frac{dt}{d\tau} \\
&=
(\vcond(\x(t)),1)\,p_t^\ccond(\x(t)) \\
&=
\big(\vcond(\x(t))p_t^\ccond(\x(t)),\,p_t^\ccond(\x(t))\big) \\
&=
\Eccondt(\xt(\tau)).
\end{align*}
\new{The reparameterization is well defined since $\pcond$ is strictly positive and $C^1$ (Assumption~1);
the resulting parameter range $(\tau_s,\tau_f)$ may be unbounded, which
Definition~\eqref{eq:if_connects_particles} allows, as it constrains only the limits at its endpoints.}
Therefore $\xt(\tau)$ is a field line of $\Eccondt$ connecting $\xt_0$ to $\condcolor{\xt_T}$.

\new{The same construction applies to every curve of $C$. Moreover, for every $s\in(0,T)$ and every Borel
set $B\subseteq\real^\dim$,
\[
\int_B \Eccondt(\x,s)_t\,d\x
=p_s^\ccond(B)
=\eta\big(e_s^{-1}(B)\big)
=\eta\big(e_s^{-1}(B)\cap C\big),
\]
the last equality because $\eta(C)=1$. The connecting field lines therefore carry the entire flux
through every slice $\{t=s\}$, and the remaining ones carry none, which is the zero-flux qualification
in~\eqref{eq:if_connects_particles}. These slices are the transversal hypersurfaces used in
Appendix~\ref{proof_3}.}

\textit{Divergence-free.}
Using
$\Eccondt(\x, t)_\x = \vcond(\x)\,\pcond(\x)$
and $\pcond(\x)=\Eccondt(\x, t)_t$,
we obtain 
\begin{align*}
    \nabla_{(\x,t)}\cdot \Eccondt(\x,t) &= \partial_t \Eccondt(\x,t)_t + \nabla_\x\cdot \Eccondt(\x,t)_\x\\ &= \partial_t \pcond(\x) + \nabla_\x\cdot\big(\vcond(\x)\,\pcond(\x)\big) = 0, 
\end{align*}
where the last equality is the continuity equation for the Conditional Flow.
\neww{This identity holds pointwise for $t\in(0,T)$.}

\textit{Fixed total flux and $\Phi_0=1$.}
\neww{Let $\varphi\in C_c^\infty(\real^\dim\times(-\infty,T))$.
Integrating the continuity equation by parts on $(\varepsilon,T)$ and
passing to the limit using Assumption~2 and the weak endpoint condition gives
\[
-\int_0^T\!\!\int_{\real^\dim}
\Eccondt\cdot\nabla_{(\x,t)}\varphi\,d\x\,dt
=\lim_{\varepsilon\to0^+}\int_{\real^\dim}
p_\varepsilon^\ccond(\x)\varphi(\x,\varepsilon)\,d\x
=\varphi(\x_0,0).
\]
Thus the zero extension satisfies
$\nabla\cdot\Eccondt=\delta_{\xt_0}$ in $\{t<T\}$ distributionally.
For any bounded, piecewise smooth domain $M$ with $\xt_0\in M$ and
$\overline M\subset\{t<T\}$, the weak flux is therefore
\[
\int_{\partial M}\Eccondt\cdot d\mathbf S
=(\nabla\cdot\Eccondt)(M)=1.
\]
\condcolor{In the two-sided case, the same calculation at $t=T$ gives
$\nabla\cdot\Eccondt=\delta_{\xt_0}-\delta_{\xt_T}$ on $\real^{\dim+1}$;
the restriction $\overline M\subset\{t<T\}$ is then unnecessary if
$\xt_T\notin\overline M$.}
Hence $\Phi_0=1$, independently of $\ccond$.}

This proves that $\Eccondt$ is a forward-only Interaction Field with unit total flux.

% ------------------------------------------------------------
\item \textbf{Recovering the global field $\E(\x, t)$.}

By the generalized superposition principle,
\[
\E(\x,t)=\int \E^\ccond(\x,t)\, d\cjoint(\ccond).
\]
Taking components and using the definition \eqref{eq:cfm_to_ifm} gives
\[
\E(\x,t)_t
=
\int p_t^\ccond(\x)\, d\cjoint(\ccond)
=
p_t(\x),
\]
where the last equality is exactly the marginalization formula for $p_t$.
Similarly,
\[
\E(\x,t)_\x
=
\int \vcond(\x)\,\pcond(\x)\, d\cjoint(\ccond).
\]
By the definition of the CFM global velocity $\v_t(\x)$ (conditional expectation given $x_t=\x$),
\[
\v_t(\x)
=
\frac{\int \vcond(\x)\,\pcond(\x)\, d\cjoint(\ccond)}{\int \pcond(\x)\, d\cjoint(\ccond)}
=
\frac{\E(\x,t)_\x}{p_t(\x)}.
\]
Therefore $\E(\x,t)_\x=\v_t(\x)p_t(\x)$, and hence
\[
\E(\x,t)=\big(\v_t(\x)\,p_t(\x),\,p_t(\x)\big).
\]

% ------------------------------------------------------------
\item \textbf{Field-line dynamics coincide with the CFM dynamics.}

For a forward-only global field, the data-space dynamics obtained by reparameterizing field lines by $t$ is
\[
\frac{d\x}{dt}=\frac{\E(\x,t)_\x}{\E(\x,t)_t},
\]
Using Item \textbf{2},
\[
\frac{\E(\x,t)_\x}{\E(\x,t)_t}
=
\frac{\v_t(\x)p_t(\x)}{p_t(\x)}
=
\v_t(\x),
\]
so the induced dynamics coincide with the \CFM{} generative ODE:
\[
\frac{d\x}{dt}=\v_t(\x)=\frac{\E(\x,t)_\x}{\E(\x,t)_t}.
\]
\end{enumerate}
\qed

\begin{remark}[Multiplicity of characteristics]
\label{rem:many_characteristics}
\new{The connecting property~\eqref{eq:if_connects_particles} does not single out one curve, and in general
the connecting family is infinite. Consider the two-sided stochastic interpolant of
\S\ref{sec_extra:two_sided_stochastic_interpolant}, with
$\vcond(\x)=\Idot{t}+({\sdot{t}}/{\s{t}})\big(\x-\I{t}\big)$.
The substitution $\mathbf{y}=(\x-\I{t})/\s{t}$ turns the characteristic ODE into $\dot{\mathbf{y}}=0$, so
its solutions form the family
\[
\x(t)=\I{t}+\s{t}\,\mathbf{y},
\qquad
\mathbf{y}\in\real^\dim .
\]
Since $\s{0}=\s{T}=0$, \emph{every} member of this family satisfies $\x(0)=\x_0$ and
$\condcolor{\x(T)=\x_T}$. The whole continuum of characteristics therefore connects the same pair of
endpoints, which is exactly what~\eqref{eq:if_connects_particles} requires. This is the field-theoretic
analogue of a point charge emitting a continuum of field lines. In this example the connecting property
holds for every curve, while in general the superposition principle yields it for $\eta$-almost every
curve, as required by~\eqref{eq:if_connects_particles}.}
\end{remark}

% ##########

% \begin{mytheorem}{Proposition~\ref{thm:flux_conservation_across_slices}}
% \label{proof:flux_conservation_across_slices}
% \textnormal{(Flux Conservation across Slices).}
% The scalar $\Phi_t^\cond$ in~\eqref{eq:flux_across_slice} is the flux through the slice
% $\{t=\mathrm{const}\}$ and coincides with the total flux~\eqref{eq:ifm_total_flux} $\Phi_0$
% of the forward-only Interaction Field.
% \end{mytheorem}

\subsection{Assumptions for Proposition~\ref{thm:flux_conservation_across_slices} and Theorem~\ref{thm:ifm_to_cfm}}
\label{assumptions_2}

Assume $\mathbf{E}^{\ccond}(\mathbf{x},t)$ is a forward-only Interaction Field
with total flux $\Phi_0$, satisfying:

\begin{enumerate}
    \item $\E^\ccond \in C^1(\mathbb{R}^D \times (0,T))$, $ $ $\E^\ccond_t (\x, t) > 0$ for $t \in (0, T)$ and $\E^\ccond_t (\x, t) = 0$ for $t < 0$;
    \item The Field decays sufficiently at infinity so that integrals converge ($\int_{\mathbb{R}^D} | \E^\ccond_t (\x, t)| d\x < \infty$) and the flux through infinite lateral surfaces vanishes:
    \[
\lim_{R\to\infty}\int_{\partial B_R\times[t_1,t_2]}\Eccondt (\x, t) \cdot d\mathbf S=0;
\]
    \item \new{\textit{(Interior completeness.)} For every $s \in (0,T)$ and every $\x \in \real^\dim$, the maximal
      existence interval of the solution of $\dot{\x} = \v_r^\ccond(\x)$, $\x(s) = \x$, with
      $\vcond := \E^\ccond_\x/\E^\ccond_t$, is all of $(0,T)$.}
\end{enumerate}

\neww{We use the boundary convention of Appendix~\ref{assumptions_1}.}

\new{Assumption~1 already makes $\vcond$ of class $C^1$ on $\real^\dim\times(0,T)$, being a quotient of
$C^1$ functions with nonvanishing denominator. Hence $\vcond$ is locally Lipschitz in $\x$, uniformly
in $t$ on compact subsets of $(0,T)\times\real^\dim$, so characteristics are unique and depend on
initial conditions in a $C^1$ manner, and no separate Lipschitz assumption is required.
This uniqueness holds only up to the maximal existence time, which is why completeness is stated
separately in Assumption~3. Together with the connectivity condition of
\S\ref{sec:interaction_field_matching}, which is part of the definition of an Interaction Field rather
than an additional requirement, these three assumptions are what the proof in
Appendix~\ref{proof_3} uses; the endpoint behaviour of $\E^\ccond_t$ is derived there and is not
assumed.}

\begin{remark}[Sufficient criterion for Assumption~3]
\label{rem:completeness_criterion}
\new{Assumption~3 holds whenever $\vcond$ grows at most linearly on compact time intervals: if
$\|\v_t^\ccond(\x)\| \le C_{a,b}\,(1+\|\x\|)$ on $[a,b]\times\real^\dim$ for every
$[a,b]\subset(0,T)$, then Gr\"onwall's inequality gives
$1+\|\x(t)\| \le (1+\|\x(s)\|)\,e^{C_{a,b}|t-s|}$, so no solution blows up inside $(0,T)$.
This criterion gives completeness only, and not the endpoint limits.
For the two-sided stochastic interpolant of
\S\ref{sec_extra:two_sided_stochastic_interpolant} both are available in closed form: the solution
through $(s,\x)$ is
\[
\gamma^{s,\x}(t) = \I{t} + \frac{\s{t}}{\s{s}}\big(\x - \I{s}\big),
\]
which is defined on all of $(0,T)$ and, since $\s{0}=\s{T}=0$, converges to $\I{0}=\x_0$ as
$t\to 0^+$ and to $\I{T}=\condcolor{\x_T}$ as $t\to T^-$.}
\end{remark}

\subsection{Proof of Proposition~\ref{thm:flux_conservation_across_slices}}
\label{proof_2}

\new{Here we use Assumptions 1 and 2 above.}  Fix any $t^*\in(0,T)$. We show that the slice flux equals the total flux:
\begin{equation}\label{appeq:total_flux}
\int_{\real^\dim}\Eccondt(\x,t^*)_t\,d\x=\Phi_0.
\end{equation}

First, for $R>0$, define the cylinder
\[
M_R:=B_R(\x_0)\times[-t^*,t^*],
\qquad
B_R(\x_0):=\{\x\in\real^\dim:\|\x-\x_0\|_2\le R\}.
\]
Since $t^*<T$, the volume $M_R$ encloses $\xt_0=(\x_0,0)$ \new{\condcolor{but not $\xt_T=(\x_T,T)$}}.
Hence, by the defining total-flux property of a forward-only Interaction Field,
\[
\int_{\partial M_R}\Eccondt\cdot d\mathbf S=\Phi_0.
\]

Next, let's decompose the boundary into bottom, lateral, and top parts:
\[
\partial M_R
=
\underbrace{B_R\times\{-t^*\}}_{\text{bottom}}
\;\cup\;
\underbrace{\partial B_R\times[-t^*,t^*]}_{\text{lateral}}
\;\cup\;
\underbrace{B_R\times\{t^*\}}_{\text{top}}.
\]
Using outward normals, we obtain
\begin{equation}\label{appeq:total_flux_decomp}
\int_{\partial M_R}\Eccondt\cdot d\mathbf S
=
-\int_{B_R}\Eccondt(\x,-t^*)_t\,d\x
+\int_{\partial B_R\times[-t^*,t^*]}\Eccondt\cdot d\mathbf S
+\int_{B_R}\Eccondt(\x,t^*)_t\,d\x .
\end{equation}

The bottom term vanishes \new{for every $R$ because, by the forward-only convention recorded in
Assumption~1, $\Eccondt(\x,t)_t=0$ for $t<0$}:
\[
\int_{B_R}\Eccondt(\x,-t^*)_t\,d\x=0.
\]
\new{By Assumption 2,} the field has vanishing lateral flux at
infinity~\citep{manukhov2025interaction}:
\[
\lim_{R\to\infty}\int_{\partial B_R\times[-t^*,t^*]}\Eccondt\cdot d\mathbf S=0,
\]
\new{and $\Eccondt(\cdot,t^*)_t$ is integrable, so the top term converges to the integral over
$\real^\dim$.}
Letting $R\to\infty$ in \eqref{appeq:total_flux_decomp} therefore yields
\[
\Phi_0
=
\lim_{R\to\infty}\int_{\partial M_R}\Eccondt\cdot d\mathbf S
=
\lim_{R\to\infty}\int_{B_R}\Eccondt(\x,t^*)_t\,d\x
=
\int_{\real^\dim}\Eccondt(\x,t^*)_t\,d\x,
\]
which proves \eqref{appeq:total_flux}.
\qed

\subsection{Proof of Theorem~\ref{thm:ifm_to_cfm}}
\label{proof_3}

We prove the three items in order.

\begin{enumerate}[itemsep=-1pt, topsep=-1pt, label=\textbf{\arabic*.}]

\item \textbf{The pair $(\vcond(\x),\pcond(\x))$ defines a Conditional Flow.}

\textit{Well-definedness.}
Since $\Eccondt$ is forward-only, $\Eccondt(\x,t)_t>0$ wherever the field is defined.
Hence $\vcond(\x)$ is well-defined and $\pcond(\x)\ge 0$.

\textit{Normalization.}
By Proposition~\ref{thm:flux_conservation_across_slices}, for every $t\in(0,T)$,
\[
\int_{\real^\dim}\Eccondt(\x,t)_t\,d\x=\Phi_0.
\]
Therefore,
\[
\int_{\real^\dim}\pcond(\x)\,d\x
=
\frac{1}{\Phi_0}\int_{\real^\dim}\Eccondt(\x,t)_t\,d\x
=1.
\]

\textit{Continuity equation.}
Using $\pcond(\x)=\Eccondt(\x, t)_t/\Phi_0$ and $\vcond(\x)\,\pcond(\x)=\Eccondt(\x, t)_\x/\Phi_0$, we obtain
\begin{align*}
\partial_t \pcond(\x) + \nabla_\x\cdot\big(\vcond(\x)\,\pcond(\x)\big)
&=
\frac{1}{\Phi_0}\Big(\partial_t \Eccondt(\x,t)_t + \nabla_\x\cdot \Eccondt(\x,t)_\x\Big)\\
&=
\frac{1}{\Phi_0}\,\nabla_{(\x,t)}\cdot \Eccondt(\x,t)
=
0,
\end{align*}
where the last equality is the divergence-free property of Interaction Fields.

\textit{Boundary conditions.}
\new{The endpoint behaviour of $\Eccondt_t$ is not assumed: we derive it from the connectivity
condition~\eqref{eq:if_connects_particles} and Assumptions 1 to 3.
Throughout, write
\[
\mu_t(d\x):=\Eccondt(\x,t)_t\,d\x,
\qquad t\in(0,T),
\]
for the slice measure, so that $\pcond=\mu_t/\Phi_0$.}

\textit{(i) Slice-wise form of the connectivity condition.}
\new{Call a hypersurface $S\subset\real^{\dim+1}$ transversal if $\Eccondt\cdot\n>0$ along it, and equip it
with the flux measure $\Eccondt\cdot\n\,d\mathcal{H}^\dim$.
Since the field is forward-only, $\Eccondt_t>0$, so every field line meets a slice $\{t=s\}$ at most
once, and a line connecting $\xt_0$ and $\condcolor{\xt_T}$ meets every intermediate slice exactly
once. The slice is thus transversal, and its flux measure is $\mu_s$.
Let $A_s\subset\real^\dim$ be the set of points whose maximal field line, unique by Assumption 1, does
not connect $\xt_0$ \condcolor{and $\xt_T$}; it is a Borel set, by the continuity of the flow
established in (ii). The flux that these lines carry through $\{t=s\}$ is
exactly $\mu_s(A_s)$, so the zero-flux qualification
in~\eqref{eq:if_connects_particles} says that $\mu_s(A_s)=0$ for every $s\in(0,T)$. Consequently, for
$\mu_s$-almost every $\x$, the field line through $(\x,s)$ connects the prescribed endpoints.}

\textit{(ii) The flow.}
\new{For $s,t\in(0,T)$ and $\x\in\real^\dim$ set $X_{s\to t}(\x):=\gamma^{s,\x}(t)$, where $\gamma^{s,\x}$ is
the maximal solution of $\dot\gamma(r)=\v_r^\ccond(\gamma(r))$, $\gamma(s)=\x$.
By Assumption 1 the field is $C^1$ with $\Eccondt_t>0$, so $\vcond$ is $C^1$ and in particular locally
Lipschitz in $\x$; characteristics are therefore unique, and Assumption 3 places $t$ in the
domain of $\gamma^{s,\x}$. Uniqueness gives $X_{t\to u}\circ X_{s\to t}=X_{s\to u}$ and
$X_{s\to s}=\mathrm{id}$, so $X_{s\to t}$ is a bijection of $\real^\dim$ with inverse $X_{t\to s}$, and
$C^1$ dependence on initial conditions makes it a $C^1$ diffeomorphism, jointly $C^1$ in $(t,\x)$.}

\textit{(iii) Transport identity.}
\new{The divergence-free property~\eqref{eq:if_conserves_flux} reads
$\partial_t\mu_t+\nabla_\x\cdot(\v_t^\ccond\mu_t)=0$ on $(0,T)\times\real^\dim$.
Fix $[a,b]\subset(0,T)$, $s\in[a,b]$ and $\varphi\in C^1_c(\real^\dim)$, and put
$\psi(r,\x):=\varphi\big(X_{r\to s}(\x)\big)$, which is defined on all of $\real^\dim$ by Assumption 3
and is $C^1$ by (ii); for $r\in[a,b]$ its supports lie in the single compact set
$K:=\{X_{s\to r}(\mathbf{y}):r\in[a,b],\,\mathbf{y}\in\operatorname{supp}\varphi\}$.
Along any solution $\gamma$, uniqueness gives $X_{r\to s}(\gamma(r))=\gamma(s)$, so
$r\mapsto\psi(r,\gamma(r))$ is constant; since a trajectory passes through every point,
$\partial_r\psi+\v_r^\ccond\!\cdot\!\nabla_\x\psi=0$.
Testing the continuity equation with $\chi(r)\,\psi(r,\x)$, $\chi\in C^1_c((a,b))$, therefore leaves
only the $\chi'$ term:
\[
0=\int_a^b\chi'(r)\,F(r)\,dr,
\qquad
F(r):=\int\psi(r,\x)\,d\mu_r(\x)=\int_K\psi(r,\x)\,\Eccondt(\x,r)_t\,d\x .
\]
The integrand is jointly continuous with support in the fixed compact set $K$, so $F$ is continuous by
dominated convergence, and, having vanishing distributional derivative, it is constant on $(a,b)$ and,
by continuity, on $[a,b]$.
Evaluating at $r=t$ and $r=s$ gives $(X_{t\to s})_\#\mu_t=\mu_s$, that is
\[
\mu_t=(X_{s\to t})_\#\mu_s,
\qquad s,t\in(0,T),
\]
since $[a,b]\subset(0,T)$ was arbitrary.}

\textit{(iv) No exchange of mass at infinity.}
\new{Every part of $\mu_t$ is the image under $X_{s\to t}$ of a part of $\mu_s$, so no mass arrives from
infinity, and a push-forward preserves total mass, so none escapes. This excludes simultaneous inflow
and outflow at infinity as well as each of them separately.}

\textit{(v) Endpoint limits.}
\new{By Proposition~\ref{thm:flux_conservation_across_slices}, which rests on Assumptions 1 and 2, each
$\mu_t$ with $t\in(0,T)$ is a finite measure of mass $\Phi_0$.
By (i), for $\mu_s$-a.e. $\x$ the field line through $(\x,s)$ connects $\xt_0$ and
$\condcolor{\xt_T}$, and by uniqueness of characteristics that line, reparameterized by $t$, is the
trajectory $r\mapsto X_{s\to r}(\x)$ of (ii). Hence $X_{s\to t}(\x)\to\x_0$ as $t\to0^+$ for
$\mu_s$-a.e. $\x$, and for every $\varphi\in C_b(\real^\dim)$,
\[
\int\varphi\,d\mu_t
=
\int\varphi\big(X_{s\to t}(\x)\big)\,d\mu_s(\x)
\;\longrightarrow\;
\Phi_0\,\varphi(\x_0),
\qquad t\to0^+,
\]
by dominated convergence, using $|\varphi|\le\|\varphi\|_\infty$ and finiteness of $\mu_s$.
\condcolor{The terminal condition follows symmetrically.}
In the sense of weak convergence of measures on $\real^\dim$, therefore,
\[
\Eccondt(\x,t)_t\,d\x \;\Rightarrow\; \Phi_0\,\delta_{\x_0}(\x)
 \ (t\to0^+),
\quad
\condcolor{\Eccondt(\x,t)_t\,d\x \;\Rightarrow\; \Phi_0\,\delta_{\x_T}(\x)
 \ (t\to T^-)}.
\]}
Dividing by $\Phi_0$ gives
\[
p^{\x_0,\x_T}_{t=0}(\x)=\delta_{\x_0}(\x),
\qquad
\condcolor{p^{\x_0,\x_T}_{t=T}(\x)=\delta_{\x_T}(\x)},
\]
which is exactly~\eqref{eq:cfm_cond_boundaries}.
Thus $(\vcond(\x),\pcond(\x))$ is a Conditional Flow.

\item \textbf{Recovering $(\v_t(\x),p_t(\x))$ from the global field $\E(\x, t)$.}

By generalized superposition~\eqref{eq:field_full},
\[
\E(\x,t) \;=\; \int \Eccondt(\x,t)\,\cjoint(\x_0,\x_T)\,d\x_0\,d\x_T.
\]
Taking the $t$-component and using $\pcond=\Eccondt_t/\Phi_0$ gives
\[
\E(\x,t)_t
=
\int \Eccondt(\x,t)_t\,\cjoint\,d\x_0\,d\x_T
=
\Phi_0\int \pcond(\x)\,\cjoint\,d\x_0\,d\x_T.
\]
By the CFM marginalization formula~\eqref{eq:cfm_marginals},
$p_t(\x)=\int \pcond(\x)\,\cjoint\,d\x_0\,d\x_T$, hence
\[
p_t(\x)=\frac{\E(\x,t)_t}{\Phi_0}.
\]
Similarly, using $\vcond(\x)\,\pcond(\x)=\Eccondt(\x, t)_\x/\Phi_0$,
\[
\E(\x,t)_\x
=
\int \Eccondt(\x,t)_\x\,\cjoint\,d\x_0\,d\x_T
=
\Phi_0\int \vcond(\x)\,\pcond(\x)\,\cjoint\,d\x_0\,d\x_T.
\]
On the other hand, by the definition of the global drift~\eqref{eq:cfm_ode_and_drift},
\begin{align*}
\v_t(\x)\,p_t(\x)
&= \mathbb E_{\ccond}\left[\vcond(\x_t)\mid \x_t=\x\right]\,p_t(\x)\\
&= p_t(\x)\int \vcond(\x)\,p_t(\x_0,\x_T\mid \x_t=\x)\,d\x_0\,\condcolor{d\x_T}\\
&= p_t(\x)\int \vcond(\x)\,
\frac{\pcond(\x)\,\cjoint(\x_0,\x_T)}{p_t(\x)}\,d\x_0\,\condcolor{d\x_T}\\
&= \int \vcond(\x)\,\pcond(\x)\,\cjoint(\x_0,\x_T)\,d\x_0\,\condcolor{d\x_T}.
\label{appeq:vtpt_mixture}
\end{align*}

Combining the last three displays yields
\[
\v_t(\x)\,p_t(\x)=\frac{\E(\x,t)_\x}{\Phi_0}.
\]
Substituting $p_t(\x)=\E(\x,t)_t/\Phi_0$ gives
\[
\v_t(\x)=\frac{\E(\x,t)_\x}{\E(\x,t)_t},
\]
proving~\eqref{eq:ifm_to_cfm_global}.

\item \textbf{Flow-based dynamics coincide with global field lines.}

Let $\tilde\x(\tau)=(\x(\tau),t(\tau))$ be a field line of the global field:
\[
\frac{d\tilde\x(\tau)}{d\tau}=\E(\tilde\x(\tau)).
\]
Since $\E$ is a nonnegative superposition of forward-only fields, $\E_t(\x,t)\ge 0$ everywhere
(and $\E_t>0$ along any nontrivial field line). Hence $\frac{dt}{d\tau}=\E_t(\tilde\x(\tau))>0$
and we may reparameterize the curve by $t$:
\[
\frac{d\x}{dt}
=
\frac{d\x/d\tau}{dt/d\tau}
=
\frac{\E(\x,t)_\x}{\E(\x,t)_t}
=
\v_t(\x),
\]
where the last equality follows from item \textbf{2}.
This is exactly the \CFM{} generative dynamics~\eqref{eq:cfm_ode_and_drift}.
\end{enumerate}
\qed

\section{Additional material for Section~\ref{sec:beyond_foifm}}
\label{app:separation_proofs}
This section collects the assumptions and proofs for the results of~\S\ref{sec:beyond_foifm}.

\subsection{Assumptions for Proposition~\ref{prop:ifm_return_orientation}}
\label{app:return_orientation}

Write $\mathbf b:=-\E$ for the generative field and assume that its
trajectories issued from $U\times\{T\}$ satisfy:

\begin{enumerate}
    \item They are unique and sweep out an open subset of
    $\real^{\dim+1}$, on which $\mathbf b$ is $C^1$ and nonzero away from the
    initial and terminal sections, with $C^1$ one-sided traces at charged
    sections;
    \item They leave $U\times\{T\}$ into $t<T$ transversally:
    $b_t(a,T^-)\ne0$ for $a \in U$;
    \item They reach the selected endpoint on $t=0$ from $t<0$
    transversally: $b_t(F(a),0^-)\ne0$ for $a \in U$;
    \item Any crossing of a plate before the selected endpoint lies outside
    the charged support, so that $\mathbf b$ extends $C^1$ across it.
\end{enumerate}

We call such a family a \emph{$C^1$ flow tube}. The implicit-function theorem
makes its hitting time $\tau(a)$ and transport map $F$ of class $C^1$ on $U$,
and both endpoint sections are oriented using the same data coordinates
$(x_1,\ldots,x_\dim)$. These assumptions constrain only the branch under
consideration and not the field away from it;
Lemma~\ref{lem:returning_tube_persistence} verifies them for the EFM
construction of Proposition~\ref{prop:efm_return_example}.

\subsection{Proof of Proposition~\ref{prop:ifm_return_orientation}}
\label{app:return_orientation_proof}

Let $\varphi_\tau$ denote the flow of the generative field $\mathbf b=-\E$,
and let $e_1,\ldots,e_\dim$ be the positively oriented coordinate basis
tangent to the plates. Fix $a\in U$ and let $\tau(a)$ be the hitting time of
the selected intersection, so that
\[
    \varphi_{\tau(a)}(a,T)=(F(a),0).
\]
We show that the orientation of $F$ is decided by the normal traces of
$\mathbf b$ at the two ends of the tube. The argument rests on the standard
Jacobian identity for a flow map between two transverse sections, which we
derive here for completeness:
\begin{equation}
    b_t(F(a),0^-)\,\det\nabla F(a)
    =
    b_t(a,T^-)\,\det D\varphi_{\tau(a)}.
    \label{appeq:section_map_jacobian}
\end{equation}

First, let's build a frame that joins the two sections. Let $M(a)$ be the
$(\dim+1)\times(\dim+1)$ matrix whose first $\dim$ columns are the
variations $\partial_{a_i}\bigl[\varphi_{\tau(a)}(a,T)\bigr]$ and whose last
column is $\mathbf b(F(a),0^-)$. We evaluate $\det M(a)$ in two ways.

\textit{At the terminal section.}
Since $\varphi_{\tau(a)}(a,T)=(F(a),0)$ for every $a\in U$, the first $\dim$
columns are $(\partial_iF(a),0)$, so expansion along the last row gives
\[
    \det M(a)
    =
    b_t(F(a),0^-)\,\det\nabla F(a).
\]

\textit{Along the flow.}
Differentiating $\varphi_{\tau(a)}(a,T)$ with respect to $a$ produces the
fixed-time variations $D\varphi_{\tau(a)}e_i$ and an additional multiple of
$\mathbf b(F(a),0^-)$ coming from the derivative of $\tau(a)$; being
multiples of the last column, they leave the determinant unchanged.
Moreover, invariance of a vector field under its own flow gives
\[
    D\varphi_{\tau(a)}\,\mathbf b(a,T^-)
    =
    \mathbf b(F(a),0^-),
\]
so the last column is the image of $\mathbf b(a,T^-)$ as well. Hence
\[
    \det M(a)
    =
    \det D\varphi_{\tau(a)}\cdot
    \det\bigl(e_1,\ldots,e_\dim,\mathbf b(a,T^-)\bigr)
    =
    b_t(a,T^-)\,\det D\varphi_{\tau(a)}.
\]
Comparing the two expressions proves~\eqref{appeq:section_map_jacobian}.

Next, let's check that the flow factor is positive. By item \textbf{1}, the
field $\mathbf b$ is $C^1$ along the tube, including the one-sided traces at
its two ends and the crossing in between, so $\varphi_\tau$ is a
diffeomorphism onto its image for every $\tau\in[0,\tau(a)]$. Therefore
$\tau\mapsto\det D\varphi_\tau$ is continuous, never vanishes, and equals
$1$ at $\tau=0$, whence $\det D\varphi_{\tau(a)}>0$.

It remains to read off the two traces. Since the tube leaves $U\times\{T\}$
into $t<T$, we have $b_t(a,T^-)<0$; since it reaches the selected endpoint
from $t<0$, we have $b_t(F(a),0^-)>0$. The traces have opposite signs,
so~\eqref{appeq:section_map_jacobian} gives $\det\nabla F(a)<0$ for every
$a\in U$.

The same computation applies to a branch reaching the target plate from
$t>0$: it replaces $b_t(F(a),0^-)$ by $b_t(F(a),0^+)<0$ and gives
$\det\nabla F(a)>0$. The reversal is thus caused by the return from $t<0$,
and not by nonmonotone motion inside the slab.
\qed

\subsection{Assumptions for Proposition~\ref{prop:efm_return_example}}
\label{app:efm_assumptions}

We use the standard EFM convention in $\real^{D+1}$:

\begin{enumerate}
    \item For a positive constant $c_D$, the Coulomb kernel is
    $K(y):=c_D\,y/\|y\|_2^{D+1}$, and the global field of target and source
    densities $\pi_0$ and $\pi_T$, carried by the plates $t=0$ and $t=T$
    respectively, is
    \begin{align}
        \E_{\pi_0,\pi_T}(x,t)
        &:={}
        \int_{\real^D}K\big((x,t)-(y,0)\big)\pi_0(y)\,dy
        \nonumber\\
        &\hspace{1.4em}
        -\int_{\real^D}K\big((x,t)-(y,T)\big)\pi_T(y)\,dy;
        \label{appeq:efm_global_field}
    \end{align}

    \item The first plate carries positive charge and the second carries
    negative charge, so generative trajectories follow
    $\mathbf b_{\pi_0,\pi_T}:=-\E_{\pi_0,\pi_T}$. This is the setting
    of~\citet{kolesov2025field,manukhov2025interaction} in the notation
    of~\S\ref{sec:interaction_field_matching}, with $z$ renamed $t$ and $L$
    renamed $T$: their field is our $\E$, their direction of motion is our
    $\mathbf b$, and the one-sided normal traces satisfy
    \[
        E_z(x,s^\pm)=\E_t(x,s^\pm)=-b_t(x,s^\pm),
        \qquad s\in\{0,T\};
    \]

    \item Trajectories follow the stochastic map
    of~\citet[Appendix~A.2, Eqs.~(25)--(27)]{manukhov2025interaction} in
    these coordinates, including its convention of initializing a trajectory
    arbitrarily close to a plate.
\end{enumerate}

\subsection{Proof of Proposition~\ref{prop:efm_return_example}}
\label{app:efm_proof}

We first record the one-sided traces of a Coulomb layer and then construct
densities whose field carries a returning tube.

\begin{lemma}[One-sided traces of a Coulomb layer]
\label{lem:coulomb_layer_jump}
Let $\sigma\in C_c^\infty(\real^D)$ be a positive surface density on the
plate $t=s$, and let
\[
    \E_\sigma(x,t)
    =c_D\int_{\real^D}
    \frac{(x-y,t-s)}{
    \bigl(\|x-y\|_2^2+(t-s)^2\bigr)^{(D+1)/2}}
    \sigma(y)\,dy.
\]
Then $\E_\sigma$ is smooth off the plate and has smooth one-sided traces.
Its normal component satisfies
\begin{align}
    (E_\sigma)_t(x,s^+)&=\kappa_D\sigma(x),
    &
    (E_\sigma)_t(x,s^-)&=-\kappa_D\sigma(x),
    \label{appeq:coulomb_jump}
\end{align}
where
\[
    \kappa_D
    :=c_D\int_{\real^D}
    \frac{du}{(1+\|u\|_2^2)^{(D+1)/2}}>0.
\]
The tangential component has the same principal-value trace from both
sides. Consequently, if a field $H$ is smooth across the plate, then adding
the layer $\sigma$ gives normal traces
$H_t(x,s)\pm\kappa_D\sigma(x)$; for a negative layer $-\sigma$, the signs
are reversed.
\end{lemma}

\begin{proof}
Writing $h=t-s$ and changing variables $y=x+|h|u$ in the normal component
gives
\[
    (E_\sigma)_t(x,s+h)
    =c_D\operatorname{sgn}(h)
    \int_{\real^D}
    \frac{\sigma(x+|h|u)}{(1+\|u\|_2^2)^{(D+1)/2}}\,du.
\]
Dominated convergence proves~\eqref{appeq:coulomb_jump}. The tangential
statement is the standard principal-value limit of the odd Coulomb kernel;
for a smooth compactly supported density this singular integral is smooth.
\end{proof}

\begin{lemma}[Persistence of a returning dipole tube]
\label{lem:returning_tube_persistence}
For every $D\ge1$, there exist a radial density
$\rho\in C_c^\infty(\real^D)$, a point $a_\star\in\real^D\setminus\{0\}$,
and, for every sufficiently small $\varepsilon>0$, smooth probability
densities
\[
    \pi_0^\varepsilon(x)=\rho_\varepsilon(x),
    \qquad
    \pi_T^\varepsilon(x)
    =(1-\delta_\varepsilon)\rho_\varepsilon(x)
    +\delta_\varepsilon\rho_\varepsilon(x-a_\star),
\]
where $\rho_\varepsilon(x)=\varepsilon^{-D}\rho(x/\varepsilon)$ and
$\delta_\varepsilon=\varepsilon^{D+2}$, with the following property. The
corresponding EFM field has a nonempty open set $U_\varepsilon$ satisfying
\[
    \overline{U_\varepsilon}
    \subset
    a_\star+\varepsilon
    \operatorname{int}(\operatorname{supp}\rho)
\]
such that every generative trajectory issued from $U_\varepsilon$ leaves
the source plate into the slab, crosses $t=0$ outside
$\operatorname{supp}\pi_0^\varepsilon$, enters $t<0$, and returns to
$t=0$ inside $\operatorname{int}(\operatorname{supp}\pi_0^\varepsilon)$.
These trajectories form a $C^1$ flow tube in the sense of
Appendix~\ref{app:return_orientation}. On $U_\varepsilon$ and its terminal
footprint $V_\varepsilon$, respectively, the one-sided traces satisfy
\[
    b_t(a,T^-)<0,
    \qquad
    b_t(a,T^+)<0,
    \qquad
    b_t(x,0^-)>0>b_t(x,0^+).
\]
\end{lemma}

\begin{proof}
We construct a transverse returning tube for two point charges and then
smooth the charges, treating in turn the middle of the tube, its departure
from the source plate, its passage through the annulus around the target
support, and its capture by the target plate.

\textit{A reference dipole.}
Put a positive unit charge at $q_+=(0,0)$ and a negative unit charge at
$q_-=(0,T)$.  Away from these two points, their field is
\begin{align}
    \E_\star(x,t)
    =c_D\left[
        \frac{(x,t)}{(\|x\|_2^2+t^2)^{(D+1)/2}}
        -
        \frac{(x,t-T)}{(\|x\|_2^2+(t-T)^2)^{(D+1)/2}}
    \right].
    \label{appeq:point_dipole}
\end{align}
This field is rotationally symmetric in $x$, and its field lines lie in
two-dimensional meridional half-planes. It has no zero away from the two
charges: equality of the two terms in~\eqref{appeq:point_dipole} would imply
\[
    \frac{y-q_+}{\|y-q_+\|_2^{D+1}}
    =
    \frac{y-q_-}{\|y-q_-\|_2^{D+1}},
\]
which is impossible because the radial map
$z\mapsto z/\|z\|_2^{D+1}$ is injective. Its potential $\psi_\star$, defined
by $\E_\star=-\nabla\psi_\star$, is the superposition of
$\pm g(\|y-q_\pm\|_2)$ with
\begin{equation}
    g(r)
    :=
    \begin{cases}
        -c_1\log r, & D=1,\\[2pt]
        \dfrac{c_D}{D-1}\,r^{-(D-1)}, & D\ge2,
    \end{cases}
    \label{appeq:coulomb_potential}
\end{equation}
so that $g$ is strictly decreasing in every dimension.

The positive charge emits positive flux into the lower half-space. Since
the two charges have zero net charge,
$\|\E_\star(y)\|=O(\|y\|^{-D-1})$ at infinity, so its absolute flux through
the sphere of radius $R$ is $O(R^{-1})$. Call a field line \emph{escaping}
if it meets spheres of arbitrarily large radius about the origin; a line
that is not escaping stays in a bounded set. The flux carried by a tube of
field lines is constant because $\nabla\cdot\E_\star=0$ away from the
charges, so a tube of fixed positive flux cannot consist of escaping lines.
Every angular neighborhood of the lower axis carries a fixed positive flux,
and therefore contains a line that stays in a bounded set.

Let $\gamma$ be such a line, emitted into $t<0$ and
traced in the direction of the field. Since $\E_\star=-\nabla\psi_\star$, the
potential is strictly decreasing along $\gamma$, at rate
$-\|\E_\star\|_2^2<0$. Its
$\omega$-limit set is nonempty and invariant, and $\psi_\star$ is constant on
it; hence it consists of zeros of $\E_\star$ or of the two charges. There are
no zeros away from $q_\pm$, and $q_+$ is excluded because
$\psi_\star\to+\infty$ there whereas $\psi_\star$ decreases along $\gamma$.
Thus $\gamma$ terminates at $q_-$ and, in particular, crosses $t=0$ on its
way. Returning field lines therefore accumulate at the lower axial line.

Reflection in the midplane, $R(x,t)=(x,T-t)$, satisfies
$\E_\star(Ry)=-DR\,\E_\star(y)$ and hence maps unoriented field lines to
unoriented field lines. A connecting line crosses the fixed plane
$t=T/2$; uniqueness of integral curves makes that line invariant under
$R$. Its off-axis crossings of the two plates are therefore paired. For
$x\ne0$,
\[
    (E_\star)_t(x,0)
    =(E_\star)_t(x,T)
    =c_D\frac{T}{(\|x\|_2^2+T^2)^{(D+1)/2}}>0,
\]
so both crossings are transverse. Since the normal component has only the
upward sign at either plate, the segment between these crossings remains in
the open slab. We choose below a non-axial returning line close to
the lower axis and denote its consecutive crossings by $(x_\star,0)$ and
$(a_\star,T)$; being non-axial, it has $x_\star\ne0$ and $a_\star\ne0$. Traced in the generative
direction $\mathbf b_\star=-\E_\star$, it starts at $(a_\star,T)$,
stays in $0<t<T$ until $(x_\star,0)$, enters $t<0$, and returns to $q_+$.
A sufficiently small tube around it has the same transverse crossings.

\textit{Smooth charges and the middle of the tube.}
Choose a radially symmetric function $\rho\in C_c^\infty(\real^D)$,
supported in the unit ball, positive in its interior, strictly decreasing
in radius for $0<\|x\|_2<1$, and normalized by $\int\rho=1$. Fix
$0<r<1$ with $\overline{B_r}\subset\{\rho>0\}$. We choose a large fixed $L$,
a small fixed $r_0>0$, and then the reference tube as specified by the
annular estimates below. Set
\[
    \rho_\varepsilon(x)
    :=\varepsilon^{-D}\rho(x/\varepsilon)
\]
and couple the two small parameters by
\[
    \delta_\varepsilon:=\varepsilon^{D+2}.
\]
For sufficiently small $\varepsilon>0$, define
\begin{align}
    \pi_0^{\varepsilon}(x)
    &:={}
    \rho_\varepsilon(x),
    \\
    \pi_T^{\varepsilon}(x)
    &:={}
    (1-\delta_\varepsilon)\rho_\varepsilon(x)
    +\delta_\varepsilon\rho_\varepsilon(x-a_\star).
    \label{appeq:efm_smooth_densities}
\end{align}
Both densities are smooth, compactly supported probability densities. For
small $\varepsilon$, the two source components have disjoint supports.

Fix a compact sub-tube $K$ around the middle part of the reference tube. We
truncate it inside a fixed neighborhood of $(a_\star,T)$ and at the sphere
$\|(x,t)\|_2=r_0$ around $q_+$, while retaining its first crossing
$(x_\star,0)$. Standard convolution estimates away from the three shrinking
supports give
\begin{equation}
    \left\|
    \E_{\pi_0^\varepsilon,\pi_T^\varepsilon}
    -\E_\star
    \right\|_{C^1(K)}\longrightarrow0.
    \label{appeq:middle_c1_convergence}
\end{equation}
Continuous dependence of ODE solutions therefore preserves the open tube,
the order of its crossings, and their transversality on $K$. In particular,
its first crossing of $t=0$ remains near $x_\star$ and outside
$\operatorname{supp}\pi_0^\varepsilon$. The field is smooth across the
plate there, so the EFM stopping probability at this crossing is zero.

\textit{Extension to the source plate.}
After taking one-sided traces, the field and first derivative generated by
the satellite bump obey
\[
    \|E_{\rm sat}\|=O(\delta_\varepsilon\varepsilon^{-D})
    =O(\varepsilon^2),
    \qquad
    \|DE_{\rm sat}\|=O(\delta_\varepsilon\varepsilon^{-D-1})
    =O(\varepsilon).
\]
Thus they are a $C^1$-small one-sided perturbation of the reference field in
a fixed neighborhood of $(a_\star,T)$. Since
$(E_\star)_t(a_\star,T)>0$, Lemma~\ref{lem:coulomb_layer_jump} and continuity
give a nonempty open set
\[
    \overline{U_\varepsilon}\subset
    a_\star+\varepsilon\operatorname{int}(\operatorname{supp}\rho)
\]
and $c_s>0$ such that
\[
    \E_t(a,T^\pm)\ge c_s,
    \qquad
    b_t(a,T^\pm)\le-c_s,
    \qquad a\in U_\varepsilon.
\]
The generative branch from each $a\in U_\varepsilon$ therefore leaves into
the slab and joins the perturbed middle tube.

\textit{Passage through the annulus.}
We bridge the fixed middle tube and the $O(\varepsilon)$ neighborhood of
the target support. Write $y=(x,t)=\varrho\omega$, where
$\varrho=\|y\|_2$ and $\omega\in S^D$, and decompose the generative field in
$B_{r_0}(q_+)$ as
\[
    \mathbf b(y)=-c_D\varrho^{-D}\omega+\mathbf p(y).
\]
Taylor expansion of the Coulomb kernel under convolution with
$\rho_\varepsilon$, whose first moment vanishes by radial symmetry, together
with the $O(1)$ bound on the field of the source plate near $q_+$, gives, for
$L\varepsilon\le\varrho\le r_0$,
\begin{equation}
    e(\varrho)
    :=\sup_{\|y\|_2=\varrho}
      \frac{\varrho^D\|\mathbf p(y)\|_2}{c_D}
    \le
    C_1\left(\frac{\varepsilon}{\varrho}\right)^2+C_2\varrho^D,
    \label{appeq:annular_multipole}
\end{equation}
where $C_1,C_2$ are independent of $\varepsilon$, $L\ge2$, and
$r_0\le T/2$ (and may be chosen uniformly for reference lines in a fixed
small cone about the lower axis). If $e\le\tfrac12$, then
$\mathbf b\cdot\omega\le-\tfrac12c_D\varrho^{-D}<0$, so $\varrho$
decreases strictly along trajectories, and
\[
    \left\|\frac{d\omega}{d\varrho}\right\|_2
    =
    \frac{\|(\Id-\omega\omega^\top)\mathbf b\|_2}
         {\varrho|\mathbf b\cdot\omega|}
    \le \frac{2e(\varrho)}{\varrho}.
\]
Thus, between radii $r_0$ and $L\varepsilon$, the direction $\omega$ changes
by at most
\[
    2\int_{L\varepsilon}^{r_0}e(\varrho)\,\frac{d\varrho}{\varrho}
    \le C_1L^{-2}+\frac{2C_2}{D}r_0^D.
\]
To make the choices above explicit, choose $L$ large enough that
$C_1L^{-2}<\tfrac14\arctan(r/L)$ and set
$\Theta:=\tfrac14\arctan(r/L)$. Then choose $r_0$ so that
$2C_2r_0^D/D<\Theta$ and $e\le\tfrac12$ on the annulus, and finally the
reference tube so that its angle with the lower axis at radius $r_0$ is less
than $\Theta$; such a line exists because the returning lines of $\E_\star$
accumulate at the lower axial line. For sufficiently small $\varepsilon$, the perturbed tube
enters $B_{r_0}(q_+)$ within angle $2\Theta$ of the lower axis and therefore
stays within angle $4\Theta=\arctan(r/L)$ of it throughout the annulus. In
particular, it stays in $t<0$ and crosses the section
\begin{equation}
    \{(x,t):t=-L\varepsilon,\ \|x\|_2<r\varepsilon\}.
    \label{appeq:target_entry_section}
\end{equation}

\textit{Capture by the target layer.}
Put $(x,t)=\varepsilon(\xi,s)$ and multiply the generative field by
$\varepsilon^D$. On compact subsets of $s\le0$, using the one-sided trace
at $s=0$,
\begin{equation}
    \varepsilon^D
    \mathbf b_{\pi_0^\varepsilon,\pi_T^\varepsilon}
      (\varepsilon\xi,\varepsilon s)
    \longrightarrow \mathbf b_\rho(\xi,s)
    \quad\text{in }C^1,
    \label{appeq:target_rescaling}
\end{equation}
where $\mathbf b_\rho$ is the reversed field of the single positive layer
$\rho$ at $s=0$. The target-layer term rescales exactly, whereas the two
source components vanish after multiplication by $\varepsilon^D$ because
their supports remain a positive distance away. For $s<0$,
\[
    (b_\rho)_s(\xi,s)
    =c_D(-s)\int_{\real^D}
      \frac{\rho(\eta)\,d\eta}
      {(\|\xi-\eta\|_2^2+s^2)^{(D+1)/2}}>0.
\]
The potential of the layer $\rho$ at a fixed height $s\le0$ is
$\psi_\rho(\xi,s)=\int_{\real^D}\rho(\eta)G_s(\xi-\eta)\,d\eta$ with
$G_s(u):=g\bigl((\|u\|_2^2+s^2)^{1/2}\bigr)$
and $g$ as in~\eqref{appeq:coulomb_potential}; thus $G_s$ is radial and
strictly decreasing in $\|u\|_2$. Write $\rho(\eta)=\hat\rho(\|\eta\|_2)$,
fix $r>0$, and take $\xi=re_1$ by rotational symmetry. Integrating by parts
in $\eta_1$ and pairing $\eta$ with its reflection
$\eta^*:=(-\eta_1,\eta_2,\ldots,\eta_\dim)$ gives
\[
    \partial_r\psi_\rho(re_1,s)
    =
    \int_{\{\eta_1>0\}}
    \hat\rho'(\|\eta\|_2)\,\frac{\eta_1}{\|\eta\|_2}
    \bigl[G_s(re_1-\eta)-G_s(re_1-\eta^*)\bigr]
    \,d\eta ,
\]
which is valid up to $s=0$, where $G_s$ is only weakly singular, because the
derivative falls on $\rho$. For $\eta_1>0$ one has
$\|re_1-\eta\|_2<\|re_1-\eta^*\|_2$, so the bracket is positive, while
$\hat\rho'\le0$ and $\hat\rho'<0$ on a set of positive measure because
$\hat\rho$ is strictly decreasing on $(0,1)$. Hence
$\partial_r\psi_\rho<0$ for every $r>0$ and $s\le0$, and since
$\mathbf b_\rho$ is the gradient of $\psi_\rho$,
\[
    (b_\rho)_\xi(\xi,s)\cdot\xi
    =\|\xi\|_2\,\partial_r\psi_\rho(\|\xi\|_2,s)<0
    \qquad (\xi\ne0,\ s\le0),
\]
where the one-sided trace is used at $s=0$. Consider the vertical cylinder
\[
    \mathcal C_L
    :=\{(\xi,s):\|\xi\|_2<r,\ -L<s<0\}.
\]
On its compact lateral boundary and closure, respectively, there are
constants $c_1,c_0>0$ such that
\[
    (b_\rho)_\xi(\xi,s)\cdot\xi\le-c_1,
    \qquad
    (b_\rho)_s(\xi,s)\ge c_0.
\]
For the second inequality, use positivity for $s<0$ and the jump trace
$(b_\rho)_s(\xi,0^-)=\kappa_D\rho(\xi)>0$ on $\overline{B_r}$.
Hence the limiting field points strictly inward on the lateral boundary,
and every trajectory entering through the bottom remains in the cylinder
and reaches $s=0$ in finite flow time. By~\eqref{appeq:target_rescaling},
the same strict inequalities hold for the rescaled full field when
$\varepsilon$ is small. Together with~\eqref{appeq:target_entry_section},
this carries the perturbed tube to $t=0$ inside a compact subset of
$\operatorname{int}(\operatorname{supp}\pi_0^\varepsilon)$.

Let $V_\varepsilon$ be its terminal footprint. The target-layer traces have
order $\varepsilon^{-D}$ there, while the field of the distant source
components is $O(1)$. Lemma~\ref{lem:coulomb_layer_jump} therefore gives,
uniformly on $V_\varepsilon$,
\[
    b_t(x,0^-)>0>b_t(x,0^+)
\]
for sufficiently small $\varepsilon$. The terminal crossing is transverse,
and the resulting trajectories form the claimed $C^1$ flow tube.
\end{proof}

We can now prove the proposition.

First, let's fix the geometry. Choose the densities and $U=U_\varepsilon$
from Lemma~\ref{lem:returning_tube_persistence}; the trajectories issued
from $U$ form a $C^1$ flow tube satisfying the assumptions of
Appendix~\ref{app:return_orientation}.

Next, let's evaluate the probabilities of the stochastic
map~\citep[Appendix~A.2, Eqs.~(25)--(27)]{manukhov2025interaction} in the
coordinates of the setup.
At the source plate, Lemma~\ref{lem:returning_tube_persistence} gives
$b_t(a,T^\pm)<0$: both one-sided generative traces point into the slab, so
no branch leaves into $t>T$ and the branch entering $t<T$ is selected with
probability $\mu=1$.
At the first target crossing, which lies outside
$\operatorname{supp}\pi_0^\varepsilon$, the field is continuous, so the
one-sided normal components agree and the stopping probability is $\nu=0$.
At the second crossing, $b_t(x,0^-)>0>b_t(x,0^+)$, so the one-sided normal
components have opposite signs and $\nu=1$. Hence a trajectory started in
$U$ follows the returning tube and terminates at its second intersection
with probability one. The main and satellite source supports are
disjoint for small $\varepsilon$ because $a_\star\ne0$, and hence
\[
    \pi_T^\varepsilon(U)
    =\delta_\varepsilon
      \int_U
      \rho_\varepsilon(x-a_\star)\,dx>0,
\]
so the returning event has positive probability.

It remains to orient the resulting transport map $F$. The traces of
Lemma~\ref{lem:returning_tube_persistence}, $b_t(a,T^-)<0$ and
$b_t(F(a),0^-)>0$, are exactly the transversality conditions of
Appendix~\ref{app:return_orientation}, so
Proposition~\ref{prop:ifm_return_orientation} gives $\det\nabla F(a)<0$ for
every $a\in U$. In particular, the reversal is not an artifact of a singular
branch: the field is smooth and nonzero along the tube away from the two
charged plates, and $F$ is a local $C^1$ diffeomorphism on $U$.
\qed

\section{Empirical Validations}
\label{app:experiments}

\neww{We first describe the toy-data setup and then examine
the three takeaways of~\S\ref{sec:duality_takeaways}:
flow--field equivalence, the probabilistic interpretation
of fields, and field-informed volume coverage.}

\subsection{Setup for the toy-data validation}
\label{app:empirical_validation_setup}

We describe the common setup shared by the three verifications of
Appendices~\ref{app:duality_verification}--\ref{app:volume_coverage_ablation},
which support the takeaways of~\S\ref{sec:duality_takeaways}. All of them are
performed on the two-dimensional toy dataset shown in
Figure~\ref{fig:toy_gaussians_dataset}.

We use two two-sided stochastic
interpolants~(\S\ref{sec_extra:two_sided_stochastic_interpolant}) and set $T=1$
throughout. For $i\in\{1,2\}$ and $t\in[0,1]$, define
\[
    \x_t^i
    =
    \mathcal{I}^i(t,\ccond)
    +
    \sigma^i(t)\,\e,
    \qquad
    \e \sim \mathcal{N}(0,\Id).
\]
Equivalently,
\[
    p_{i,t}^{\ccond}(\x)
    =
    \mathcal{N}
    \left(
        \x
        \mid
        \mathcal{I}^i(t,\ccond),
        \sigma^i(t)^2 \Id
    \right).
\]

The first interpolant uses a linear path and the second a trigonometric one,
\[
    \mathcal{I}^1(t,\ccond)
    =
    (1-t)\x_0+t\x_T,
    \qquad
    \mathcal{I}^2(t,\ccond)
    =
    \cos(\pi t/2)\,\x_0+\sin(\pi t/2)\,\x_T,
\]
while both use the same noise schedule
$\sigma^1(t)=\sigma^2(t)=\sigmaval\sqrt{t(1-t)}$.

Each interpolant induces a conditional flow
$(\v_{i,t}^{\ccond},p_{i,t}^{\ccond})$, where
\[
    \v_{i,t}^{\ccond}(\x)
    =
    \partial_t \mathcal{I}^i(t,\ccond)
    +
    \frac{\dot{\sigma}^i(t)}{\sigma^i(t)}
    \left(
        \x-\mathcal{I}^i(t,\ccond)
    \right).
\]
By the flow-to-field construction, it also induces a forward-only conditional
interaction field
\[
    \E_i^{\ccond}(\x,t)
    =
    \left(
        \v_{i,t}^{\ccond}(\x)\,p_{i,t}^{\ccond}(\x),
        \;
        p_{i,t}^{\ccond}(\x)
    \right),
\]
with the global field obtained by the generalized superposition
principle~\eqref{eq:field_full},
\[
    \E_i(\x,t)
    =
    \mathbb{E}_{\ccond}
    \left[
        \E_i^{\ccond}(\x,t)
    \right].
\]
These two interpolants define the matched CFM--IFM pairs used throughout
Appendices~\ref{app:duality_verification}--\ref{app:volume_coverage_ablation}.

Unless stated otherwise, training points are drawn as $t\sim\mathcal{U}[0,T]$ and
$\x\sim p_{i,t}^{\ccond}$, so IFM is trained with the volume distribution induced
by the field itself, $p_{\mathrm{vol}} = p_{i,t}^{\ccond}$, and both models compared
in Appendix~\ref{app:duality_verification} see points from the same distribution.

\begin{figure}[htb]
    \centering
    \includegraphics[width=0.4\linewidth]{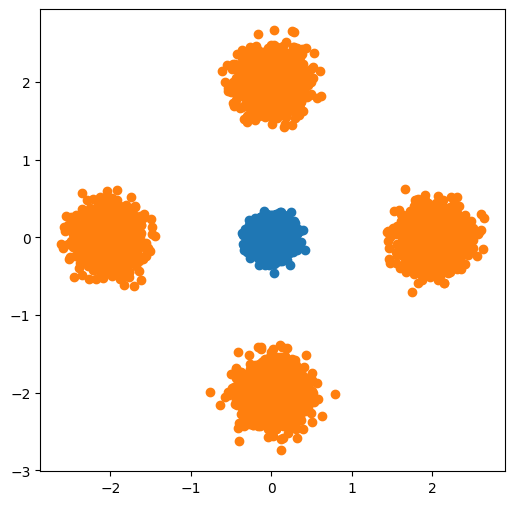}
    \caption{
        Two-dimensional toy Gaussian dataset used for the empirical validation.
        Blue points denote samples from the source distribution $\pi_T$, a
        narrow Gaussian centered near the origin. Orange points denote samples
        from the target distribution $\pi_0$, a symmetric four-component
        Gaussian mixture with modes arranged along the coordinate axes.
    }
    \label{fig:toy_gaussians_dataset}
\end{figure}

\subsection{Empirical verification of the duality}
\label{app:duality_verification}
Equations~\eqref{eq:cfm_to_ifm} and~\eqref{eq:ifm_to_cfm} make the equivalence
half of Takeaway~1 of~\S\ref{sec:duality_takeaways} directly testable: they map a
conditional flow to a forward-only interaction field and back, so the two
descriptions of one interpolant must induce the same data-space dynamics.

For each interpolant $i\in\{1,2\}$ of
Appendix~\ref{app:empirical_validation_setup} we train two separate models, each
with its own objective and its own target construction. The CFM model
learns the drift $\v_i$ from the conditional flow
$(\v_{i,t}^{\ccond},p_{i,t}^{\ccond})$ with the CFM
objective~\eqref{eq:cfm_objective}; the IFM model learns the global field $\E_i$
from the conditional field $\E_i^{\ccond}$ with the normalized-field
objective~\eqref{eq:ifm_objective}.

We then compare the data-space velocities the two models induce: the CFM drift
$\v_i$, and the drift read off the trained field~\eqref{eq:ifm_to_cfm_global},
\[
    \u_i(\x,t)
    :=
    \frac{\E_i(\x,t)_\x}{\E_i(\x,t)_t},
\]
which is a ratio of components and therefore unaffected by the normalization of
the field. If the duality holds, a matched pair $(\v_i,\u_i)$, both sides from the
same interpolant, must describe the same dynamics, while a mismatched pair, from
different interpolants, must not.

We report the mean squared error between the two drifts and the
relative error
\begin{align}\label{eq:relmse}
    \mathrm{RelMSE}(\v,\u)
    =
    \mathbb{E}
    \left[
        {\|\u(\x,t)-\v(\x,t)\|_2^2}
        \,/\,
        {\|\v(\x,t)\|_2^2}
    \right],
\end{align}
where the first listed drift of each comparison serves as the reference $\v$.
\new{Both drifts are evaluated at the same $(\x,t)$, obtained by
deterministic linear interpolations between sampled endpoints.}

Table~\ref{tab:duality_empirical_verification} confirms this prediction.
The matched MSEs are of order $10^{-7}$, six orders of magnitude below the
mismatched ones, and the matched relative errors four to five orders below:
$\v_1$ agrees with $\u_1$, and $\v_2$ agrees with $\u_2$,
whereas dynamics built from different interpolants remain far apart. The two
representations therefore coincide when they come from the same interpolant,
while different interpolants define genuinely different data-space dynamics.
Since $\u_i$ and $\v_i$ are the same drift, comparing the two fields and
comparing the two flows is the same measurement, which is why the two mismatched
rows coincide.
\begin{table}[htb]
\centering
\footnotesize
\setlength{\tabcolsep}{6pt}
\caption{
MSE and RelMSE between the CFM drift and the IFM-induced drift, for pairs built
from the same interpolant (matched) and from different interpolants (mismatched).
Here $\u_i=\E_i(\x,t)_\x/\E_i(\x,t)_t$ and RelMSE is defined
in~\eqref{eq:relmse}.
}
\begin{tabular}{llcc}
\toprule
\textbf{Type} & \textbf{Comparison}
    & \textbf{MSE} ($\downarrow$)
    & \textbf{RelMSE} ($\downarrow$) \\
\midrule
Matched
    & $\v_1$ vs. $\u_1$
    & $\mathbf{2.05\times10^{-7}}$ & $\mathbf{1.08\times10^{-6}}$ \\
Matched
    & $\v_2$ vs. $\u_2$
    & $\mathbf{2.84\times10^{-7}}$ & $\mathbf{1.34\times10^{-5}}$ \\
\midrule
Mismatched
    & $\v_1$ vs. $\v_2$
    & $4.93\times10^{-1}$ & $0.34$ \\
Mismatched
    & $\u_1$ vs. $\u_2$
    & $4.93\times10^{-1}$ & $0.34$ \\
\bottomrule
\end{tabular}
\label{tab:duality_empirical_verification}
\end{table}

\subsection{Probabilistic interpretation of a forward-only field}
\label{app:probabilistic_interpretation}
We illustrate Takeaway~2 of~\S\ref{sec:duality_takeaways}
using the first interpolant.
We evolve particles under the dynamics induced by the IFM model
trained in Appendix~\ref{app:duality_verification} and record
their empirical distributions at several times.
We compare these distributions with heatmaps of the last
component $\E_1(\x,t)_t$ of the global field, estimated by
Monte Carlo averaging the conditional fields over sampled
endpoint pairs.
As shown in Figure~\ref{fig:ifm_probabilistic_interpretation},
the transported particles concentrate in regions where this
component is large, supporting the probabilistic interpretation
predicted by the flow--field duality.

\begin{figure}[htb]
    \centering
    \includegraphics[width=\linewidth]{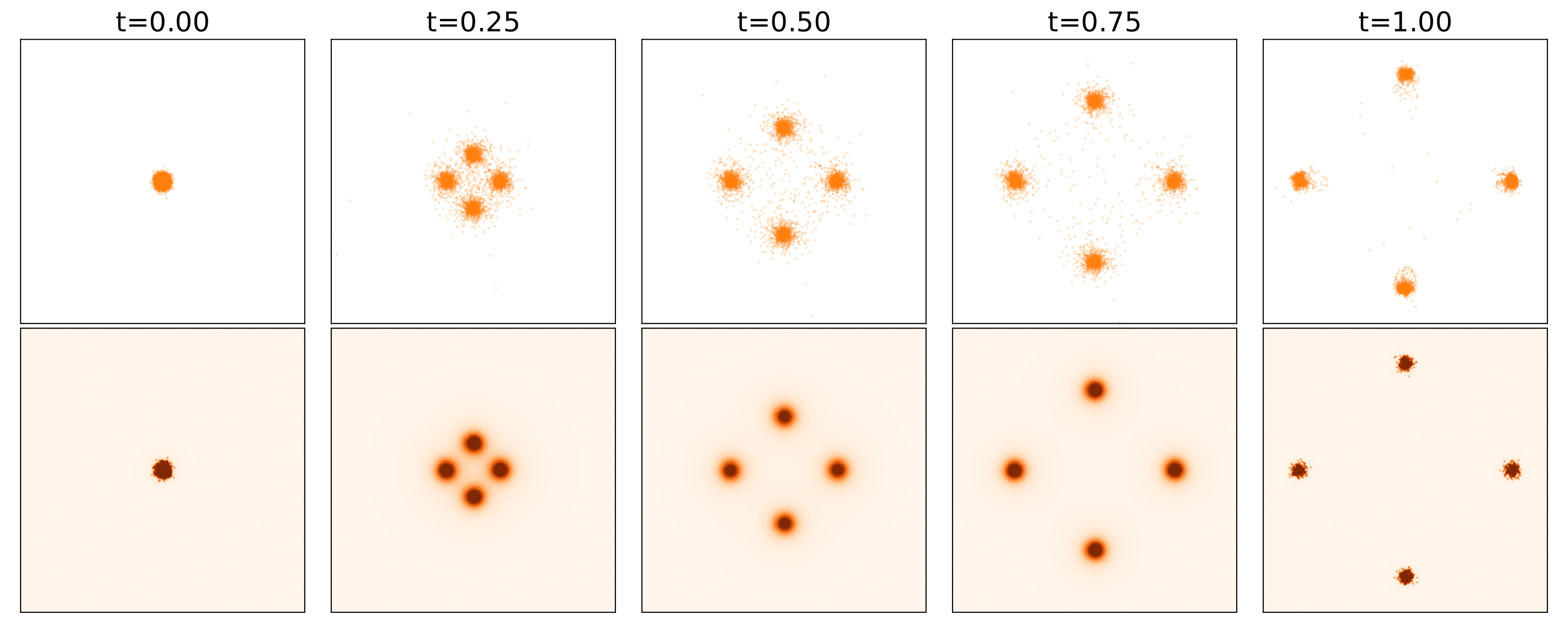}
    \caption{
        Probabilistic interpretation of a forward-only interaction field.
        Top: particles evolved under the dynamics induced by the trained
        IFM model.
        Bottom: heatmaps of the last component $\E_1(\x,t)_t$ of the
        global field, estimated by Monte Carlo averaging the conditional
        fields over sampled endpoint pairs.
    }
    \label{fig:ifm_probabilistic_interpretation}
\end{figure}

\subsection{Field-informed volume coverage}
\label{app:volume_coverage_ablation}
To assess whether the field-induced density of Takeaway~3
of~\S\ref{sec:duality_takeaways} provides a useful training distribution, we
perform a matched--mismatched ablation. For each field $\E_i$, we train the
normalized-field objective~\eqref{eq:ifm_objective} with volume distribution
$p_{\mathrm{vol}}$ set either to the matched path $p_{i,t}^{\ccond}$, the
conditional density read off the same field as
$\E_i^{\ccond}(\x,t)_t/\Phi_0$, which
reproduces~\eqref{eq:ifm_objective_with_new_volume}, or to the mismatched path
$p_{j,t}^{\ccond}$ read off the other field, $j\neq i$. Thus, the objective and
the target field are fixed, and only the volume-coverage distribution is
changed.

\begin{table}[htb]
\centering
\small
\caption{
Ablation of the volume-coverage distribution. For each field $\E_i$, the
normalized-field objective~\eqref{eq:ifm_objective} is trained with
$p_{\mathrm{vol}}$ set to the matched path $p_{i,t}^{\ccond}$ or to the mismatched
path $p_{j,t}^{\ccond}$, $j\neq i$; the last column reports the final $W_2$
distance.
\new{Reported as mean $\pm$ standard deviation over 3 repeated evaluations of each
trained model, with randomly resampled source subsets.}
}
\setlength{\tabcolsep}{5pt}
\begin{tabular}{llc}
\toprule
\textbf{Field} & \textbf{Volume distribution} & \textbf{$W_2$ ($\downarrow$)} \\
\midrule
\multirow{2}{*}{$\E_1$}
    & \textbf{matched} $p_{1,t}^{\ccond}$    & \textbf{0.25$\pm$0.04} \\
    & mismatched $p_{2,t}^{\ccond}$ & 0.58$\pm$0.03 \\
\midrule
\multirow{2}{*}{$\E_2$}
    & mismatched $p_{1,t}^{\ccond}$ & 0.81$\pm$0.16 \\
    & \textbf{matched} $p_{2,t}^{\ccond}$    & \textbf{0.38$\pm$0.04} \\
\bottomrule
\end{tabular}

\label{tab:p_vol_ablation}
\end{table}

Table~\ref{tab:p_vol_ablation} shows that matched sampling consistently
achieves a lower final $W_2$ distance. This supports the interpretation that
the $t$-component of a forward-only interaction field identifies where training
points should be placed in the augmented space, providing a principled
field-informed replacement for heuristic volume coverage.

\section{Additional Discussion and Experiments on Duality}
\label{app:additional_discussion_and_experiments}
\newcommand{\condi}[1]{{{\x_0^{#1}, \condcolor{\x_T^{#1}}}}}
\newcommand{\vcondi}[1]{{\v_t^\condi{#1}}}
\newcommand{\pcondi}[1]{{p_t^\condi{#1}}}
\newcommand{\Econdi}[1]{{\E^\condi{#1}}}
Takeaways~2 and~3 of~\S\ref{sec:duality_takeaways} are mostly about what CFM
lends IFM: an explicit marginal path, and a volume distribution that no longer has
to be guessed. The duality also runs the other way, and what IFM lends CFM is its
estimator. By Theorem~\ref{thm:cfm_to_ifm} the global drift of a conditional flow
is the ratio of the components of its dual field, and that field is estimated by
superposition over endpoint pairs~\eqref{eq:field_full}. Read back in data space,
the single-sample CFM target~\eqref{eq:cfm_objective} becomes a multi-sample one,
\begin{align}\label{eq:multisample_target}
    \hat{\v}^\batchsize_t(\x)
    :=
    \frac{\hat{\E}^\batchsize(\x,t)_\x}{\hat{\E}^\batchsize(\x,t)_t}
    =
    \frac{\sum_{i=1}^\batchsize \vcondi{i}(\x)\,\pcondi{i}(\x)}
         {\sum_{i=1}^\batchsize \pcondi{i}(\x)}
    =
    \sum_{i=1}^\batchsize \omega_i\,\vcondi{i}(\x),
\end{align}
with $\omega_i := \pcondi{i}(\x)/\sum_{k=1}^\batchsize \pcondi{k}(\x)$ and
$\x_0^i,\condcolor{\x_T^i}\sim\cjoint$. This appendix asks whether the borrowed
estimator helps conditional flows in practice.

\neww{ExFM~\citep{ryzhakov2024explicit}
derives density-weighted targets for noisy two-sided Gaussian
paths with arbitrary endpoint couplings; related estimators
appear in~\citep{bertrand2025closed,xu2023stable}.
Here, the same type of target follows from superposition
of the dual interaction fields.}

\subsection{Setup}
\label{app:multisample_setup}

\textbf{Two-sided interpolant.} Following~\citep{albergo2023stochastic}, it is
defined by $\x_t = (1-t)\,\x_0 + t\,\x_T + C\sqrt{t(1-t)}\,\varepsilon$ with
$\varepsilon\sim\mathcal{N}(0,\Id)$, where $C$ controls the injected noise
variance; since the original paper carries an extra $\sqrt{2}$, we report results
in terms of $s := C/\sqrt{2}$. Figure~\ref{fig:scale-fid} shows that larger $s$
consistently degrades sample quality, so we use $s = 0.1$ throughout. The network
regresses the conditional velocity of Table~\ref{tab:cfm_ifm_dual_constructions}
through~\eqref{eq:multisample_target}.

\begin{figure}[t]
    \centering
    \includegraphics[width=0.48\linewidth]{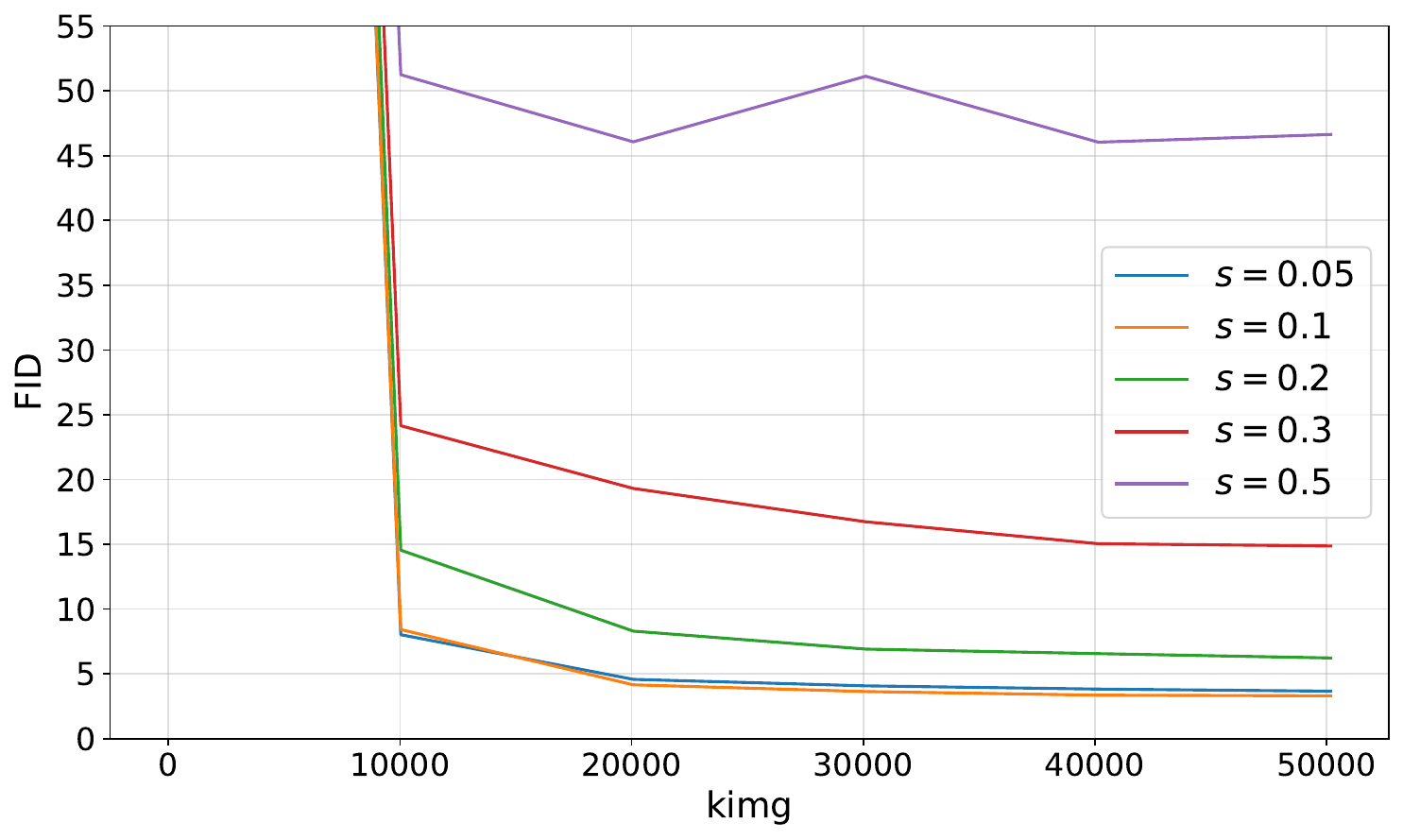}
    \caption{
        FID vs.\ training progress for the two-sided interpolant at different
        noise scales $s$.
    }
    \label{fig:scale-fid}
\end{figure}

\textbf{One-sided interpolants.} For EDM~\citep{karras2022elucidating} and
PFGM++~\citep{pfgm++} the conditional velocities are also given in
Table~\ref{tab:cfm_ifm_dual_constructions}.
\neww{Following PFGM++~\citep{pfgm++}, we parameterize the auxiliary radius as $r=\sqrt{\extradim}\,\sigma$.}
The weights
in~\eqref{eq:multisample_target} are computed from $p_\sigma(\x_\sigma\mid\x_0^i)$
for $\x_0^i\sim\pi_0$. To keep the original hyperparameters and samplers, we
convert the velocity estimate into a clean-sample one:
\begin{align}\label{eq:edm_multisample_x0}
    \x_0^\batchsize(\x_\sigma)
    &= \x_\sigma - \sigma\,\hat{\v}_\sigma^\batchsize(\x_\sigma)
    \nonumber\\
    &= \x_\sigma
       - \sigma\,
         \frac{\sum_{i=1}^{\batchsize} \v_\sigma^{\x_0^i}(\x_\sigma)\,
               p_\sigma(\x_\sigma \mid \x_0^i)}
              {\sum_{i=1}^{\batchsize} p_\sigma(\x_\sigma \mid \x_0^i)}
    \nonumber\\
    &= \x_\sigma
       - \frac{\sum_{i=1}^{\batchsize} (\x_\sigma - \x_0^i)\,
               p_\sigma(\x_\sigma \mid \x_0^i)}
              {\sum_{i=1}^{\batchsize} p_\sigma(\x_\sigma \mid \x_0^i)} ,
\end{align}
where the last equality uses $\v_\sigma^{\x_0^i}(\x_\sigma) =
(\x_\sigma-\x_0^i)/\sigma$. These models are trained to predict $\x_0$ directly,
following the original EDM and PFGM++ conventions.

\textbf{Datasets and protocol.}
We build on the official PFGM++ codebase~\citep{pfgm++}, preserving the original
architecture and training configuration as faithfully as possible; the only
notable modification is a training batch size of $B=256$ instead of $512$ for
CIFAR-10~\citep{cifar}. All image experiments are unconditional generation on
CIFAR-10 and FFHQ-64~\citep{ffhq}, trained for $100$k kimg on CIFAR-10 and
$200$k kimg on FFHQ, where one kimg is one thousand training images processed.

For the multi-sample target we use a target set of size $\batchsize \ge B$: when
$\batchsize > 1$, the target is estimated from the current batch of size $B$, on
which the loss is evaluated, together with $\batchsize - B$ additional samples
drawn at random from the dataset, so the estimate always uses a pool at least as
large as the gradient-evaluation batch.

For each FID~\citep{fid} evaluation we draw $50$k samples and compare against the
full training set, with $50$ solver steps and the Euler sampler for the two-sided
interpolant and the Heun sampler for EDM and PFGM++. All image experiments ran on
2$\times$A100 GPUs; each CIFAR-10 run took about two days and each FFHQ run about
eight.

\subsection{Results}
\label{app:multisample_results}

Table~\ref{tab:multisample_results} reports CIFAR-10 FID as a function of the
number of samples $\batchsize$ used for target estimation, and
Figure~\ref{fig:fid_converg} the corresponding training curves.

For the two-sided interpolant nothing changes: the curves are nearly
indistinguishable across all tested $\batchsize$
(Figure~\ref{fig:fid_converg}a), and the FID stays within $3.03$--$3.05$, its
best value reached at $\batchsize = 1$.

For EDM and PFGM++ the multi-sample target helps, improving both convergence
speed and final FID (Figures~\ref{fig:fid_converg}b--d), though not monotonically
in $\batchsize$: EDM is best at $\batchsize = 2048$, $2.29 \to 2.12$, while both
PFGM++ variants peak at $\batchsize = 256$, $2.40 \to 2.24$ for
$\extradim = 128$ and $2.40 \to 2.14$ for $\extradim = 2048$.
\neww{For the trained checkpoints, the observed FID gaps exceed the standard
deviations across evaluation seeds ($0.02$--$0.05$).}

\begin{table}[t]
\centering
\footnotesize
\caption{
FID ($\downarrow$)~\citep{fid} on CIFAR-10 as a function of the number of samples
$\batchsize$ used for the multi-sample target. Best scores per column are
\textbf{bolded}. Standard deviations are measured across 3 different sets of seeds
for FID evaluation.
}
\label{tab:multisample_results}
\begin{tabular}{c cccc}
\toprule
\multirow{2}{*}{$\batchsize$}
& \multicolumn{4}{c}{\textbf{FID} ($\downarrow$)} \\
\cmidrule(lr){2-5}
& \makecell[c]{\textbf{Two-sided}\\\textbf{Linear}}
& \textbf{EDM}
& \makecell[c]{\textbf{PFGM++}\\$(\extradim=128)$}
& \makecell[c]{\textbf{PFGM++}\\$(\extradim=2048)$} \\
\midrule
1
& $\mathbf{3.03 \pm 0.03}$
& $2.29 \pm 0.04$
& $2.40 \pm 0.04$
& $2.40 \pm 0.05$ \\
256
& $3.04 \pm 0.03$
& $2.21 \pm 0.03$
& $\mathbf{2.24 \pm 0.03}$
& $\mathbf{2.14 \pm 0.04}$ \\
2048
& $3.05 \pm 0.04$
& $\mathbf{2.12 \pm 0.02}$
& $2.28 \pm 0.03$
& $2.19 \pm 0.03$ \\
\bottomrule
\end{tabular}
\end{table}

\begin{figure*}[t]
    \centering
    \begin{subfigure}[t]{0.48\textwidth}
        \centering
        \includegraphics[width=\linewidth]{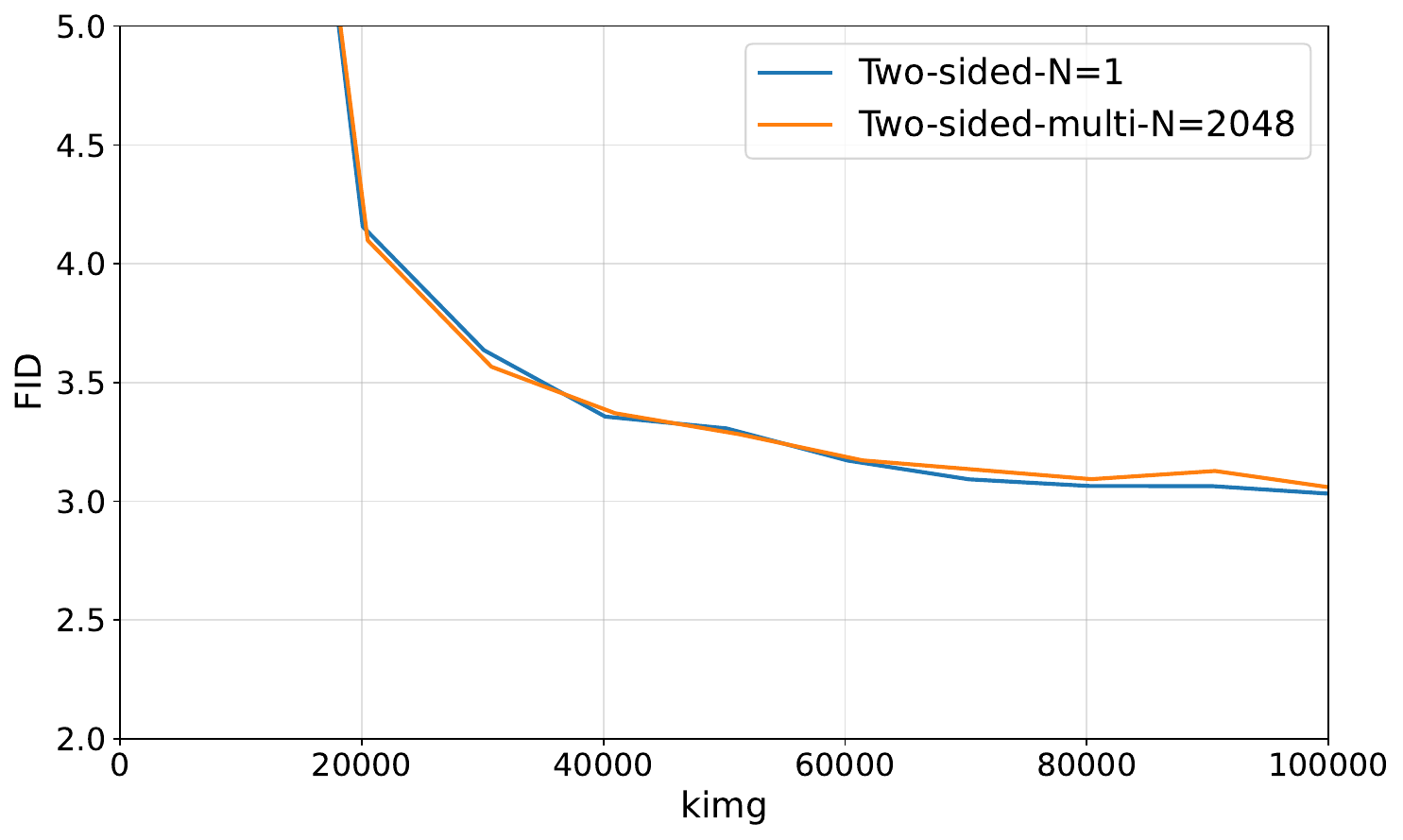}
        \caption{Two-sided interpolant ($s = 0.1$).}
    \end{subfigure}
    \hfill
    \begin{subfigure}[t]{0.48\textwidth}
        \centering
        \includegraphics[width=\linewidth]{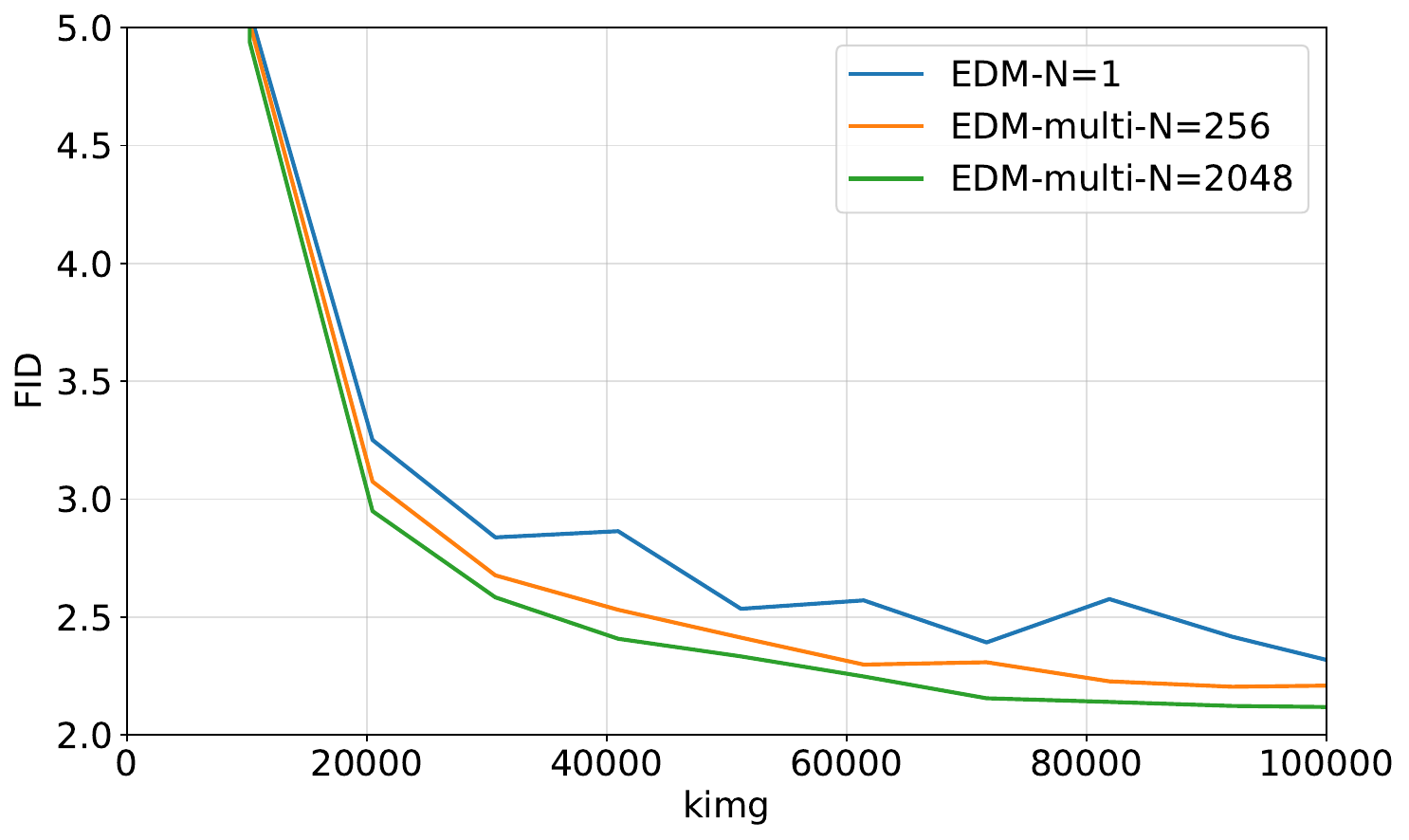}
        \caption{EDM.}
    \end{subfigure}
    \vspace{0.8em}
    \begin{subfigure}[t]{0.48\textwidth}
        \centering
        \includegraphics[width=\linewidth]{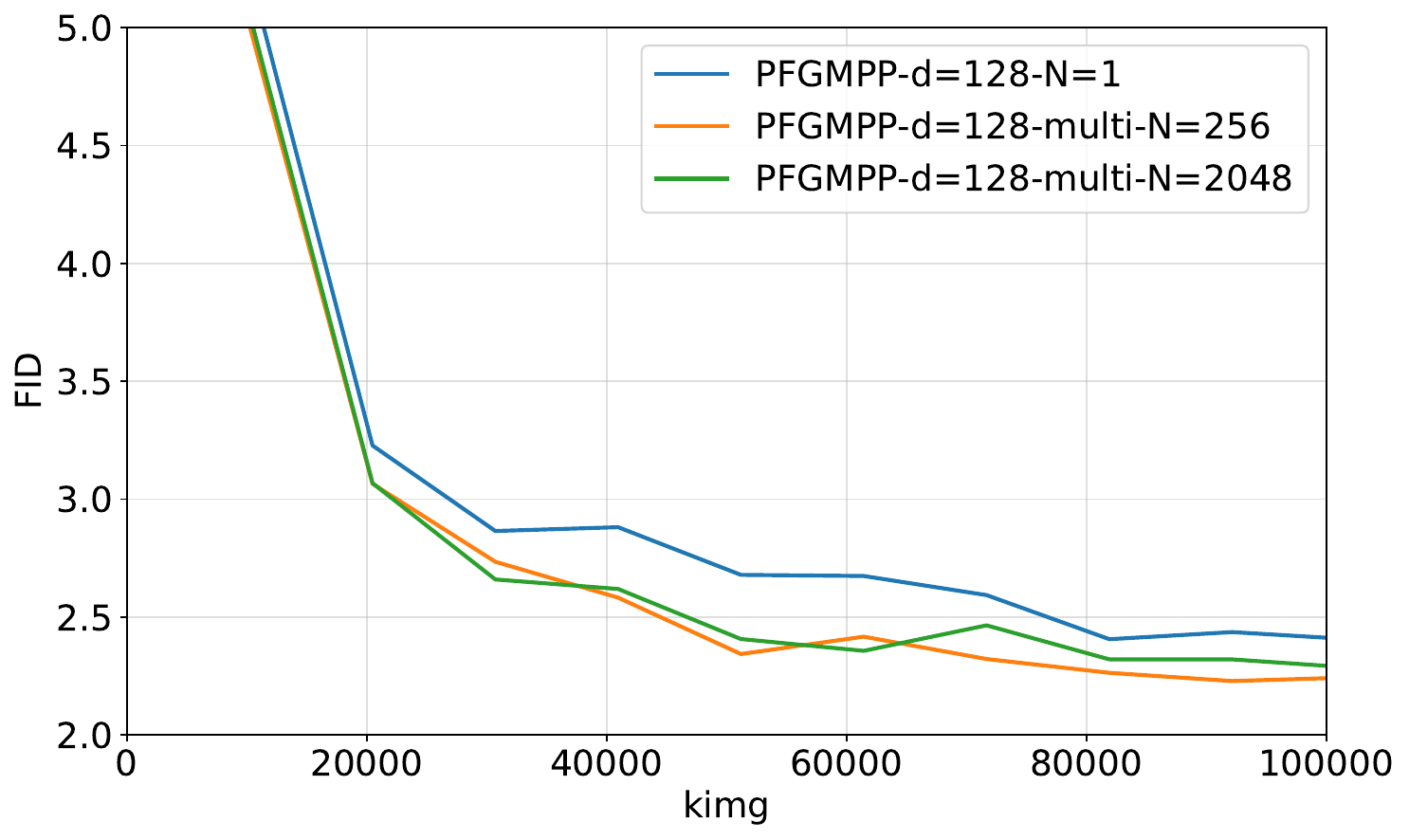}
        \caption{PFGM++ ($\extradim = 128$).}
    \end{subfigure}
    \hfill
    \begin{subfigure}[t]{0.48\textwidth}
        \centering
        \includegraphics[width=\linewidth]{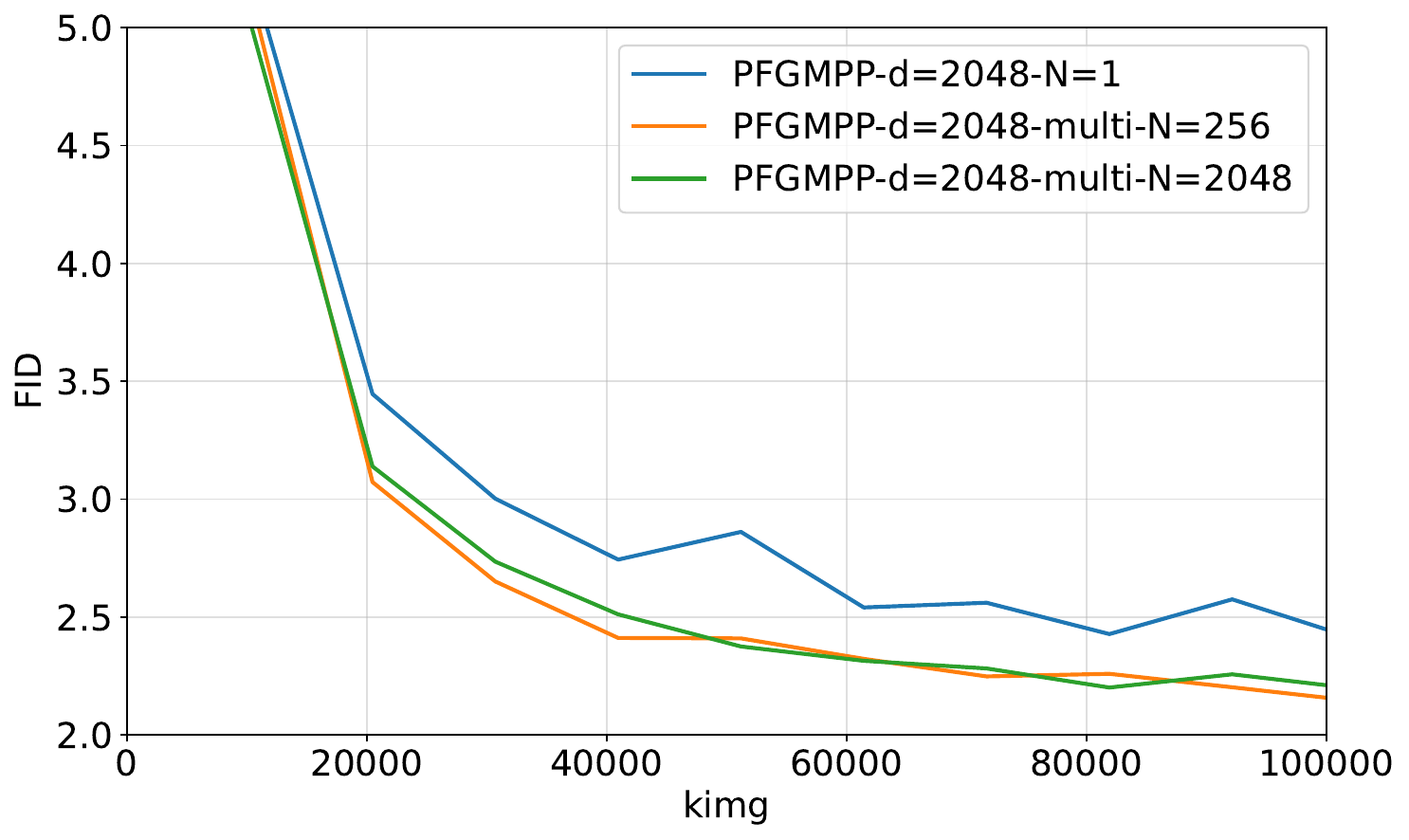}
        \caption{PFGM++ ($\extradim = 2048$).}
    \end{subfigure}
    \caption{
        Training convergence on CIFAR-10 (FID vs.\ kimg) for different generative
        methods and multi-sample target sizes $\batchsize$. One kimg = one
        thousand images processed.
    }
    \label{fig:fid_converg}
\end{figure*}

On FFHQ-64 the ordering is the same but the margin is narrower
(Figure~\ref{fig:ffhq_comparison}, Table~\ref{tab:ffhq-fid}): both methods are
best at $\batchsize = 256$, $2.62 \to 2.51$ for EDM and $2.69 \to 2.56$ for
PFGM++ with $\extradim = 128$, and by $\batchsize = 2048$ EDM is back at the single-sample level while
PFGM++ retains only a small part of the gain.
Appendix~\ref{app:estimator_degeneracy} ties this narrowing
to the resolution.

\begin{table}[t]
    \centering
    \footnotesize
    \caption{
    FID ($\downarrow$) at the final checkpoint on FFHQ-64 as a function of the
multi-sample target size $\batchsize$. Best scores per column are
\textbf{bolded}. Standard deviations are measured as in
Table~\ref{tab:multisample_results}.
    }
    \label{tab:ffhq-fid}
    \begin{tabular}{c cc}
    \toprule
    \multirow{2}{*}{$\batchsize$}
    & \multicolumn{2}{c}{\textbf{FID} ($\downarrow$)} \\
    \cmidrule(lr){2-3}
    & \textbf{EDM}
    & \makecell[c]{\textbf{PFGM++}\\$(\extradim=128)$} \\
    \midrule
    1
    & $2.62 \pm 0.04$
    & $2.69 \pm 0.02$ \\
    256
    & $\mathbf{2.51 \pm 0.05}$
    & $\mathbf{2.56 \pm 0.03}$ \\
    2048
    & $2.63 \pm 0.04$
    & $2.61 \pm 0.03$ \\
    \bottomrule
    \end{tabular}
    \end{table}
\begin{figure}[t]
    \centering
    \begin{subfigure}{0.48\linewidth}
        \centering
        \includegraphics[width=\linewidth]{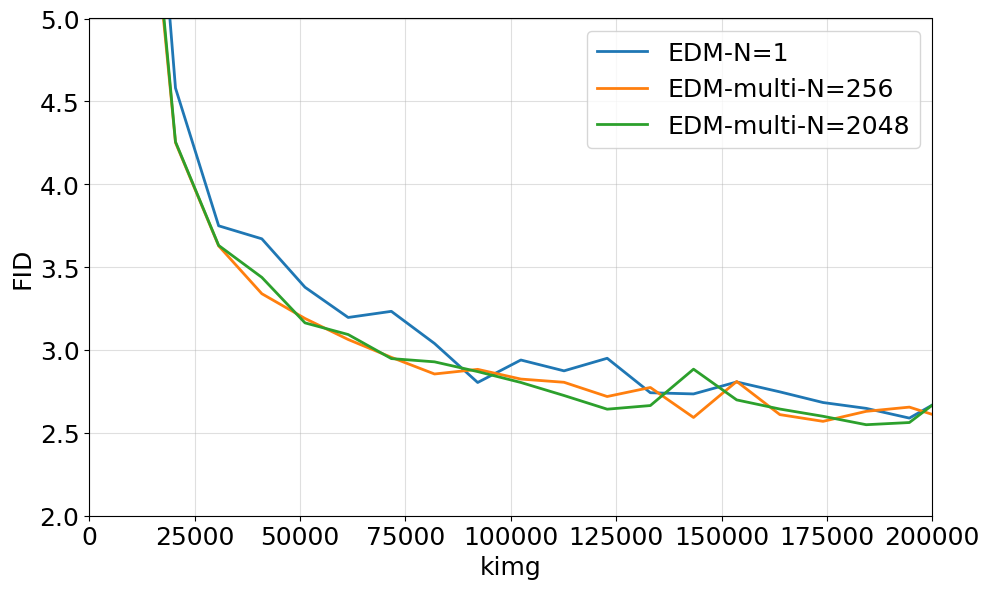}
        \caption{EDM.}
        \label{fig:ffhq_edm}
    \end{subfigure}
    \hfill
    \begin{subfigure}{0.48\linewidth}
        \centering
        \includegraphics[width=\linewidth]{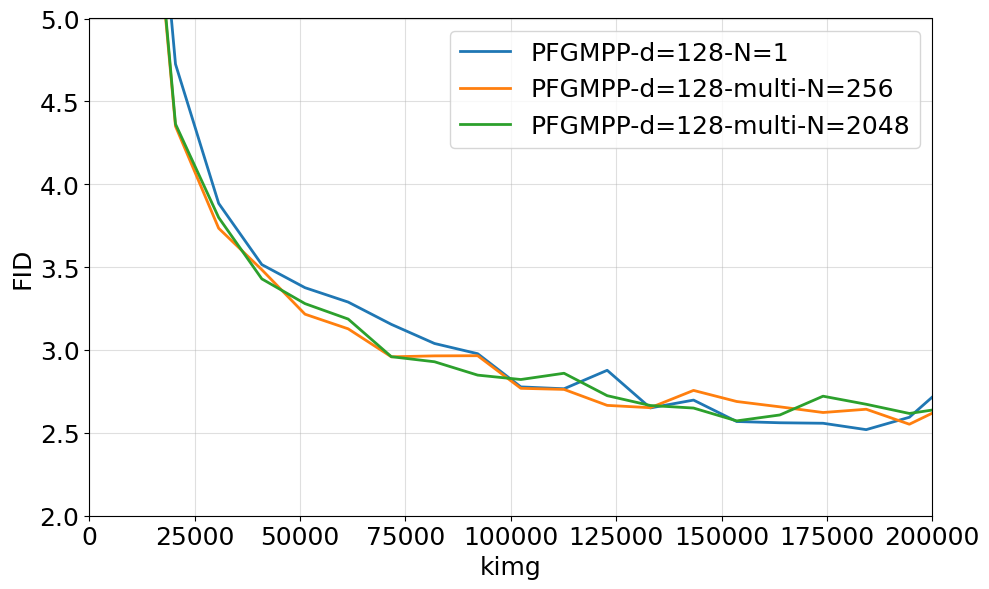}
        \caption{PFGM++ ($\extradim = 128$).}
        \label{fig:ffhq_pfgm}
    \end{subfigure}
    \caption{
        Training convergence on FFHQ-64 (FID vs.\ kimg) for the single-sample
        target and multi-sample targets with $\batchsize = 256$ and
        $\batchsize = 2048$.
    }
    \label{fig:ffhq_comparison}
\end{figure}

\subsection{Estimator degeneracy at higher resolution}
\label{app:estimator_degeneracy}

The estimator~\eqref{eq:multisample_target} is a weighted average, so it is only
as multi-sample as its weights allow. We measure their concentration by the Gini
index of the weights $\omega_i$,
\begin{equation}\label{eq:gini}
    G = 1 - \sum_{i=1}^{\batchsize} \omega_i^2 ,
\end{equation}
which is zero when one weight carries everything and grows as the weight mass
spreads out.

For the two-sided interpolant the weights are degenerate already at
$32{\times}32$: a single conditional term dominates the sum, and empirically that
term is almost always the pair $(\x_0,\condcolor{\x_T})$ used to generate the
intermediate point $\x_t \sim \pcond$, so additional samples contribute
negligibly. Removing that pair from the estimator drives the FID above $100$,
confirming that the remaining samples cannot replace it. For EDM and PFGM++ the
weights are less concentrated and spread further as $\batchsize$ grows, which is
where the gains of Appendix~\ref{app:multisample_results} come from. This also
suggests that the estimator could be improved by relevance-based sampling of
conditional pairs instead of uniform sampling.

Table~\ref{tab:gini_values_capped} shows how the concentration depends on
dimensionality: as the image resolution grows, $G$ falls by orders of magnitude,
so the multi-sample target drifts towards the single-sample one.

\begin{center}
\small
\captionof{table}{
Gini index $G$ of the weights of the multi-sample target~\eqref{eq:multisample_target} for the
EDM conditional path $p_\sigma(\x_\sigma\mid\x_0)=\mathcal{N}(\x_0,\sigma^2\Id)$, with
$\sigma$ drawn from the noise distribution proposed by EDM.
Columns vary the resolution of the CIFAR images the target samples are drawn from; rows
vary the target-set size $\batchsize$.
}
\label{tab:gini_values_capped}
\begin{tabular}{c cccc}
\toprule
\multirow{2}{*}{$\batchsize$}
& \multicolumn{4}{c}{\textbf{Resized CIFAR Resolution}} \\
\cmidrule(lr){2-5}
& \textbf{8x8}
& \textbf{16x16}
& \textbf{32x32}
& \textbf{64x64} \\
\midrule
128
& $0.214$
& $0.098$
& $0.005$
& $0.001$ \\
256
& $0.211$
& $0.112$
& $0.005$
& $0.001$ \\
512
& $0.214$
& $0.125$
& $0.006$
& $0.001$ \\
1024
& $0.216$
& $0.139$
& $0.007$
& $0.001$ \\
2048
& $0.215$
& $0.154$
& $0.009$
& $0.002$ \\
\bottomrule
\end{tabular}
\end{center}

\subsection{Conclusion}

The IFM estimator transfers: for the one-sided methods a multi-sample target
improves both convergence and final FID. Its contribution, however, shrinks as the dimensionality grows, the gain at FFHQ
$64{\times}64$ being noticeably smaller than at CIFAR-10 $32{\times}32$, and this
is exactly what the weight concentration predicts: in high dimension one
conditional term dominates, and the multi-sample estimator degenerates into the
single-sample one.

\newpage
% !TeX root = main.tex
\section*{NeurIPS Paper Checklist}

\begin{enumerate}

\item {\bf Claims}
    \item[] Question: Do the main claims made in the abstract and introduction accurately reflect the paper's contributions and scope?
    \item[] Answer: \answerYes{}.
    \item[] Justification: \neww{The main claims are supported by Sections~\ref{sec:duality}--\ref{sec:beyond_foifm}
    and the empirical validations in Appendix~\ref{app:experiments}.}


\item {\bf Limitations}
    \item[] Question: Does the paper discuss the limitations of the work performed by the authors?
    \item[] Answer: \answerYes{}.
    \item[] Justification: \neww{The theoretical scope and assumptions are stated in
    Sections~\ref{sec:duality}--\ref{sec:beyond_foifm} and
    Appendices~\ref{assumptions_1}--\ref{assumptions_2}. Practical limitations
    are discussed in Section~\ref{sec:interaction_field_matching} and
    Appendix~\ref{app:estimator_degeneracy}.}


\item {\bf Theory assumptions and proofs}
    \item[] Question: For each theoretical result, does the paper provide the full set of assumptions and a complete (and correct) proof?
    \item[] Answer: \answerYes{}.
    \item[] Justification: \neww{Assumptions and proofs are provided in
    Appendices~\ref{app:duality_proofs} and~\ref{app:separation_proofs}.
    The orientation argument underlying Corollary~\ref{cor:separation}
    is given in Section~\ref{sec:beyond_foifm}.}


    \item {\bf Experimental result reproducibility}
    \item[] Question: Does the paper fully disclose all the information needed to reproduce the main experimental results of the paper to the extent that it affects the main claims and/or conclusions of the paper (regardless of whether the code and data are provided or not)?
    \item[] Answer: \answerYes{}.
    \item[] Justification: \neww{Experimental protocols are described in
    Appendices~\ref{app:experiments} and~\ref{app:additional_discussion_and_experiments};
    architectures, optimizers and remaining hyperparameters are provided in
    the public code at \url{https://github.com/daniil-shlenskii/flow-field-matching-duality}.}


\item {\bf Open access to data and code}
    \item[] Question: Does the paper provide open access to the data and code, with sufficient instructions to faithfully reproduce the main experimental results, as described in supplemental material?
    \item[] Answer: \answerYes{}.
    \item[] Justification: \neww{The code is publicly released at
    \url{https://github.com/daniil-shlenskii/flow-field-matching-duality}.}


\item {\bf Experimental setting/details}
    \item[] Question: Does the paper specify all the training and test details (e.g., data splits, hyperparameters, how they were chosen, type of optimizer) necessary to understand the results?
    \item[] Answer: \answerYes{}.
    \item[] Justification: \neww{Training and evaluation settings are described in
    Appendices~\ref{app:empirical_validation_setup} and~\ref{app:multisample_setup},
    with architectures, optimizers and remaining hyperparameters in the released code.}


\item {\bf Experiment statistical significance}
    \item[] Question: Does the paper report error bars suitably and correctly defined or other appropriate information about the statistical significance of the experiments?
    \item[] Answer: \answerYes{} % Replace by \answerYes{}, \answerNo{}, or \answerNA{}.
    \item[] Justification: \neww{For each trained model, we report mean $\pm$ standard
    deviation over three evaluations: FID with different evaluation
    seed sets (Tables~\ref{tab:multisample_results}
    and~\ref{tab:ffhq-fid}) and $W_2$ with resampled source subsets
    (Table~\ref{tab:p_vol_ablation}).}

\item {\bf Experiments compute resources}
    \item[] Question: For each experiment, does the paper provide sufficient information on the computer resources (type of compute workers, memory, time of execution) needed to reproduce the experiments?
    \item[] Answer: \answerYes{}.
    \item[] Justification: \neww{Compute resources and training times for the image experiments
    are reported in Appendix~\ref{app:multisample_setup}.}

    
\item {\bf Code of ethics}
    \item[] Question: Does the research conducted in the paper conform, in every respect, with the NeurIPS Code of Ethics \url{https://neurips.cc/public/EthicsGuidelines}?
    \item[] Answer: \answerYes{}.
    \item[] Justification: \neww{The work is theoretical and uses public benchmark data
    (Appendices~\ref{app:empirical_validation_setup} and~\ref{app:multisample_setup}),
    with no human-subject studies or consequential deployment.}


\item {\bf Broader impacts}
    \item[] Question: Does the paper discuss both potential positive societal impacts and negative societal impacts of the work performed?
    \item[] Answer: \answerNA{}.
    \item[] Justification: \neww{The work establishes a theoretical relation between existing
    frameworks and introduces no new model, dataset or deployment; we identify
    no societal impact specific to this contribution.}

    
\item {\bf Safeguards}
    \item[] Question: Does the paper describe safeguards that have been put in place for responsible release of data or models that have a high risk for misuse (e.g., pre-trained language models, image generators, or scraped datasets)?
    \item[] Answer: \answerNA{}.
    \item[] Justification: \neww{We release experimental code, with no high-risk pretrained model
    or dataset requiring special safeguards.}


\item {\bf Licenses for existing assets}
    \item[] Question: Are the creators or original owners of assets (e.g., code, data, models), used in the paper, properly credited and are the license and terms of use explicitly mentioned and properly respected?
    \item[] Answer: \answerYes{}.
    \item[] Justification: \neww{CIFAR-10~\citep{cifar} is cited as requested and has no explicit
    license. FFHQ~\citep{ffhq} and PFGM++~\citep{pfgm++} use CC BY-NC-SA 4.0;
    individual FFHQ images retain their CC BY 2.0, CC BY-NC 2.0,
    Public Domain Mark 1.0, CC0 1.0 or U.S. Government Works licenses.
    All assets are used for non-commercial research.}

    \item[] Guidelines:
    \begin{itemize}
        \item The answer \answerNA{} means that the paper does not use existing assets.
        \item The authors should cite the original paper that produced the code package or dataset.
        \item The authors should state which version of the asset is used and, if possible, include a URL.
        \item The name of the license (e.g., CC-BY 4.0) should be included for each asset.
        \item For scraped data from a particular source (e.g., website), the copyright and terms of service of that source should be provided.
        \item If assets are released, the license, copyright information, and terms of use in the package should be provided. For popular datasets, \neww{\url{https://paperswithcode.com/datasets}} has curated licenses for some datasets. Their licensing guide can help determine the license of a dataset.
        \item For existing datasets that are re-packaged, both the original license and the license of the derived asset (if it has changed) should be provided.
        \item If this information is not available online, the authors are encouraged to reach out to the asset's creators.
    \end{itemize}

\item {\bf New assets}
    \item[] Question: Are new assets introduced in the paper well documented and is the documentation provided alongside the assets?
    \item[] Answer: \answerYes{}.
    \item[] Justification: \neww{The released code is documented in the README at
    \url{https://github.com/daniil-shlenskii/flow-field-matching-duality}.}


\item {\bf Crowdsourcing and research with human subjects}
    \item[] Question: For crowdsourcing experiments and research with human subjects, does the paper include the full text of instructions given to participants and screenshots, if applicable, as well as details about compensation (if any)? 
    \item[] Answer: \answerNA{}.
    \item[] Justification: \neww{The work involves no crowdsourcing or human-subject studies.}


\item {\bf Institutional review board (IRB) approvals or equivalent for research with human subjects}
    \item[] Question: Does the paper describe potential risks incurred by study participants, whether such risks were disclosed to the subjects, and whether Institutional Review Board (IRB) approvals (or an equivalent approval/review based on the requirements of your country or institution) were obtained?
    \item[] Answer: \answerNA{}.
    \item[] Justification: \neww{The work involves no human-subject studies requiring IRB review.}


\item {\bf Declaration of LLM usage}
    \item[] Question: Does the paper describe the usage of LLMs if it is an important, original, or non-standard component of the core methods in this research? Note that if the LLM is used only for writing, editing, or formatting purposes and does \emph{not} impact the core methodology, scientific rigor, or originality of the research, declaration is not required.
    %this research? 
    \item[] Answer: \answerNA{} % Replace by \answerYes{}, \answerNo{}, or \answerNA{}.
    \item[] Justification: \neww{LLMs are not a component of the proposed methods or experimental pipeline.}


\end{enumerate}

\end{document}